\documentclass[journal]{IEEEtran}
\ifCLASSINFOpdf
  \usepackage[pdftex]{graphicx}
\else
\fi
\usepackage[cmex10]{amsmath}
\usepackage{amsfonts}
\usepackage{amssymb}\usepackage{amsthm}\usepackage{bm}

\usepackage{amsfonts}

\usepackage{algorithmic}
\usepackage{algorithm}
\usepackage{boxedminipage}

\usepackage{xcolor}

\usepackage[normalem]{ulem}
\usepackage{verbatim}

\begin{document}
%
\title{Deep Asymmetric Networks with a Set of Node-wise Variant Activation Functions}
%
%
%

\author{Jinhyeok~Jang,
	Hyunjoong~Cho,
	Jaehong~Kim,
	Jaeyeon~Lee,
        and~Seungjoon~Yang,~\IEEEmembership{Member,~IEEE}
\thanks{J. Jang, J. Kim, and J. Lee are with Electronics and Telecommunications Research Institute (ETRI), Daejeon, Korea (e-mail:jangjh6297@gmail.com, jhkim504@etri.re.kr, leejy@etri.re.kr).}
\thanks{
H. Cho and S. Yang are with School of Electrical and Computer Engineering, Ulsan National Institute of Science and Technology
(UNIST), Ulsan, Korea (e-mail: hyeunjoong@unist.ac.kr, syang@unist.ac.kr). 
}
\thanks{
This work was supported by the ICT R\&D program of Ministry of Science and ICT / Institute for Information \& Communications Technology Promotion under Grant (2017-0-00162, Development of Human-care Robot Technology for Aging Society)}
\thanks{
This work was supported by the Ulsan National Institute of Science and Technology Free Innovation Research Fund under Grant 1.170067.01
}
}

\maketitle

\begin{abstract}
This work presents deep asymmetric networks with a set of node-wise variant activation functions. The nodes' sensitivities are affected by activation function selections such that the nodes with smaller indices become increasingly more sensitive. As a result, features learned by the nodes are sorted by the node indices in the order of their importance. Asymmetric networks not only learn input features but also the importance of those features. Nodes of lesser importance in asymmetric networks can be pruned to reduce the complexity of the networks, and the pruned networks can be retrained without incurring performance losses. We validate the feature-sorting property using both shallow and deep asymmetric networks as well as deep asymmetric networks transferred from famous networks. 
\end{abstract}

\ifCLASSOPTIONpeerreview
\begin{center} \bfseries EDICS Category: 3-BBND \end{center}
\fi
%
\IEEEpeerreviewmaketitle

\section{Introduction}
%
%
%
%

\IEEEPARstart{N}{eural} networks usually consist of neurons that have equal learning capabilities because the mathematical models of neurons are identical for all the neurons in a network. Neurons are trained by capturing the relation between their inputs and outputs. Thus, all the neurons in a network have an equal chance of learning input features. Consequently, without further inspections, one cannot tell whether a feature learned by one neuron is more or less important than features learned by other neurons. In this work, we provide network neurons with unequal feature-learning abilities; thus, some neurons learn more important features than others.

Neural networks are often trained using backpropagation \cite{rumelhart1986learning, haykin1994neural}. Errors between network outputs and target outputs are propagated backward to update the weights of nodes in previous layers of the network. The updates are proportional to both the inputs and the so-called sensitivities of the nodes. By assigning different activation functions to nodes in a layer, we allow the nodes to have different sensitivities to the same inputs. We use a set of functions parameterized by a single parameter as activation functions to assign node-wise variant sensitivities. Specifically, the slopes of the activation functions are controllable via this parameter. Thus, nodes with smaller node indices are assigned higher sensitivities. Features learned by a network with a set of node-wise variant activation functions are sorted by importance. When nodes in a trained network are removed individually, from last to first, the network accuracy gradually deteriorates at increasingly larger increments. A network with a set of node-wise variant activation function has the ability to learn not only the features that represent the inputs but also the importance of the learned features. We call deep networks containing nodes with unequal and asymmetric learning abilities ``deep asymmetric networks.''

Deep networks have achieved great successes across a wide range of fields; however, they are usually computationally expensive and memory intensive. By designing an efficient network that requires less computation but provides the same performance, one can deploy the network on a system with small computational power or perform more tasks on a system with high computational power. Many studies have investigated designing more efficient deep networks \cite{cheng2017survey, reed1993pruning}. In many approaches, the importance of nodes in a trained network are evaluated with a certain measure such as  $l_2$, $l_1$ norms, and correlation \cite{han2015learning, ishikawa1996structural, rodriguez2016regularizing}. Then the nodes showing smaller measures are pruned from the network. The performance of pruned networks depends on how effectively nonessential nodes are identified by the measure. We use the ability of asymmetric networks to learn the importance of features to design more efficient deep networks. Because the nodes in a deep asymmetric networks sort features by importance, the nodes can be pruned from least to most important to meet network computational complexity and memory requirements.

The ability to learn the importance of features is validated with both a simple network in an auto-associative setting using Gaussian data and the MNIST dataset \cite{lecun1998gradient} and deep convolutional neural networks (CNN) for object recognition using the CIFAR-10 dataset \cite{krizhevsky2009learning} and for action recognition using NTU RGB+D action recognition dataset \cite{du2015skeleton}. After the asymmetric networks are trained, we can analyze the individual contribution of each node to the reconstruction error and recognition accuracy. The experimental results show that the reconstruction error and the accuracy are increasingly influenced by the nodes with smaller indices than by those with larger indices, which indicates that the features learned by the asymmetric networks are sorted in the order of their importance. We applied the feature-sorting property to prune the deep asymmetric CNN without a loss of recognition accuracy. Using the asymmetric technique, we were also able to prune deep networks transferred from famous complex networks. To do this, we prepared VGG \cite{simonyan2014very} and ResNet \cite{he2016deep} for a facial expression recognition task. Then, we transferred the weights from the famous networks to the asymmetric networks and trained them on the CK+ dataset \cite{lucey2010extended}. The asymmetric networks pruned by the proposed procedure result in smaller (and thus more efficient) networks but exhibit no loss of accuracy. We also compared our pruning to results reported in studies using the MNIST using CIFAR-10 dataset and ResNet using ImageNet dataset.

The rest of this paper is organized as follows. Section \ref{sec:proposed} proposes the use of node-wise variant activation functions in asymmetric deep networks. The network architecture and training are introduced in Section \ref{sec:architecture}. In section \ref{sec:analysis}, the feature-sorting property of asymmetric networks is analyzed using a simple shallow network. Section \ref{sec:pruning} presents the pruning procedure used to design more efficient deep asymmetric networks. A pruning algorithm is given in Section \ref{sec:pruningalgorithm}, and a review of pruning methods is given in Section \ref{sec:pruningreview}. Experimental validations are presented in Section \ref{sec:ex} for shallow, deep, and transferred deep asymmetric networks in Sections \ref{sec:ex}. Finally, conclusions are given in Section \ref{sec:conclusion}.

\section{Deep Asymmetric Networks with a Set of Node-wise Variant Activation Functions}
\label{sec:proposed}

\subsection{Network Architecture}
\label{sec:architecture}

\begin{figure}[!t]
	\centering		
	
	\begin{minipage}{\linewidth}		
		\centering
		{\includegraphics[trim = 0 0 0 0, clip, width=\linewidth]{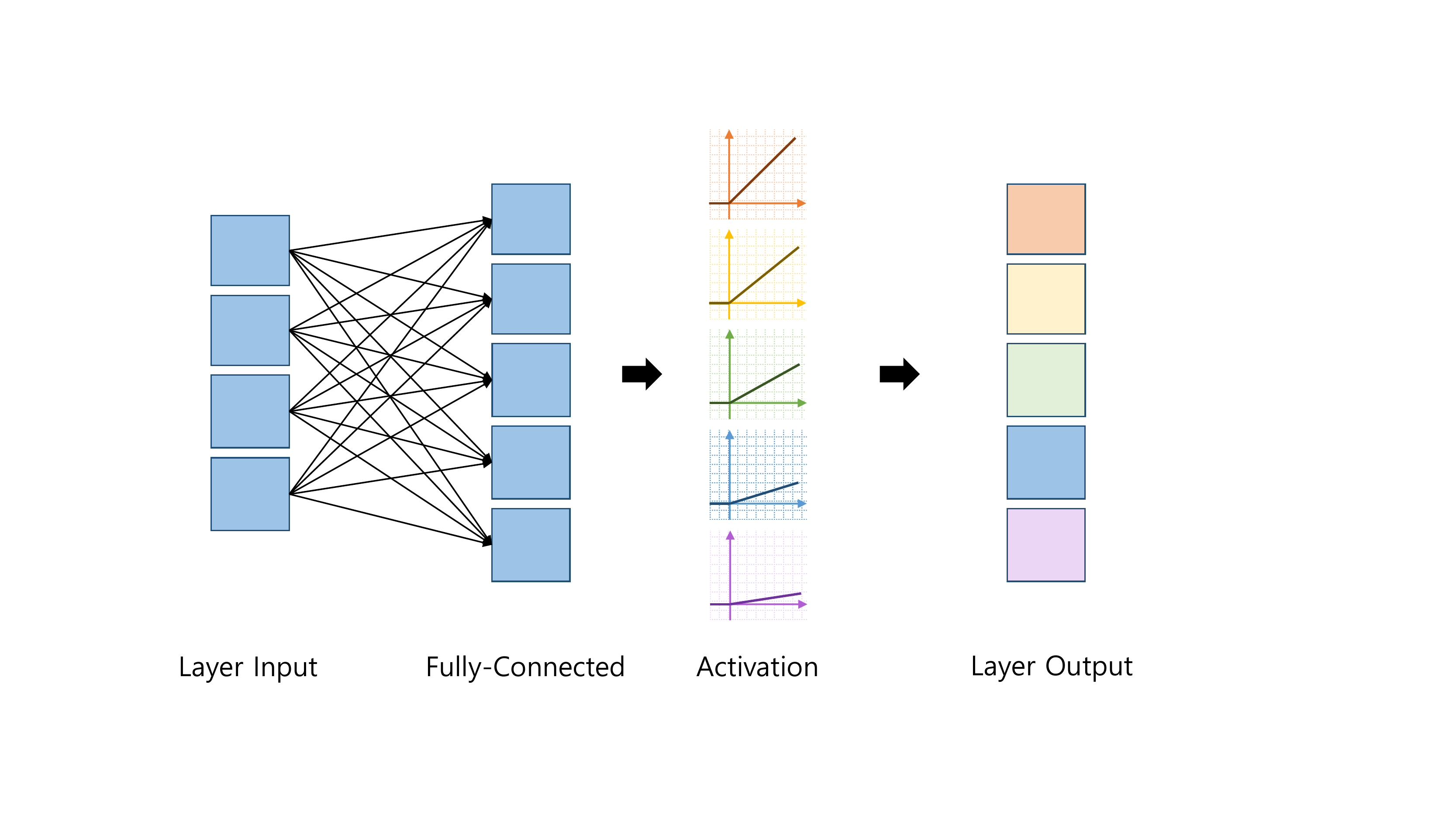}}%
		
		{\footnotesize (a)}
	\end{minipage}%
		
	\begin{minipage}{\linewidth}		
		\centering
		{\includegraphics[trim = 0 0 0 0, clip, width=\linewidth]{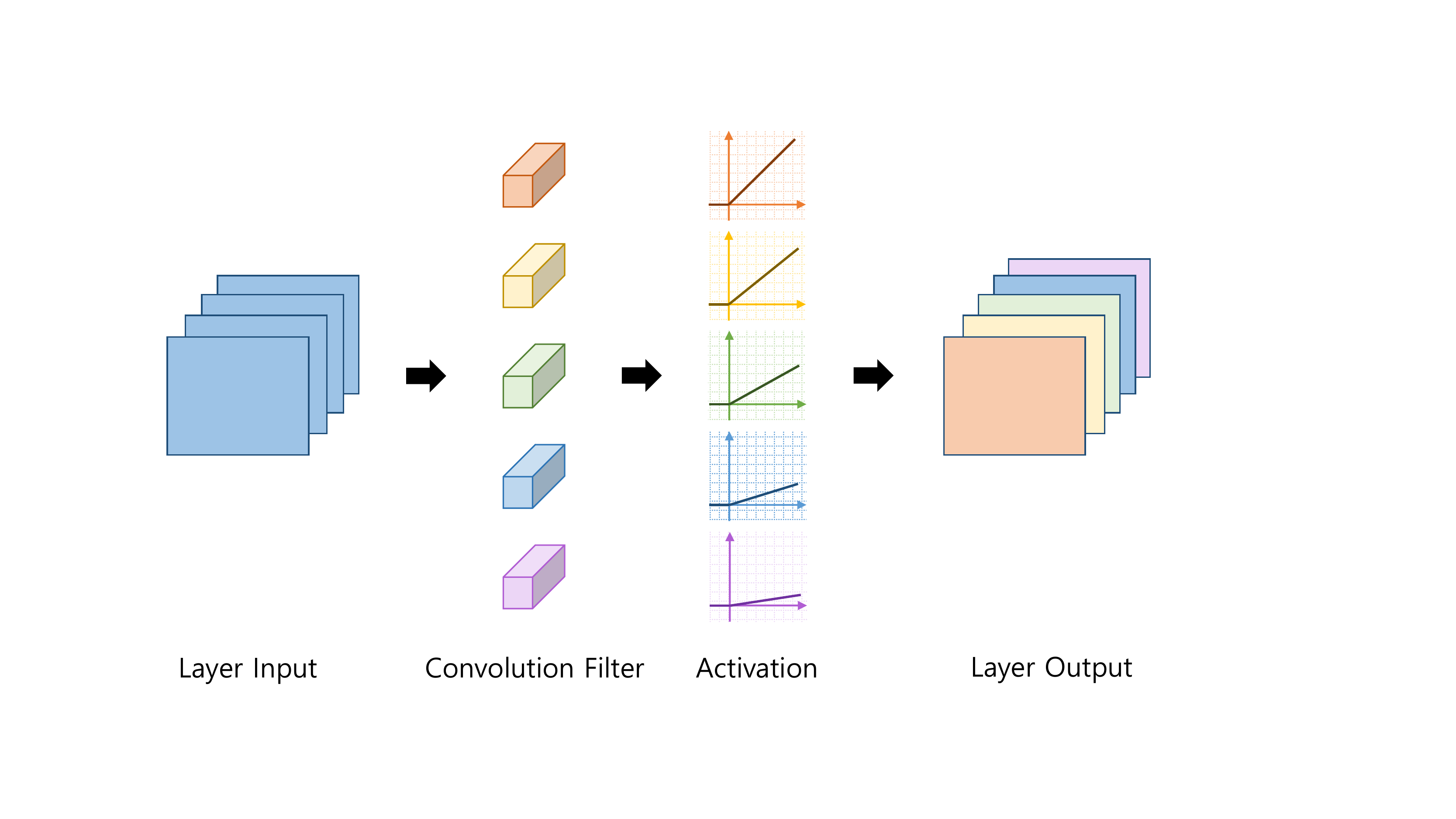}}%
		
		{\footnotesize (b)}
	\end{minipage}%
	\caption{A schematic of a layer in the proposed asymmetric network with a set of node-wise variant activation functions,
	(a) fully connected layer, and 
	(b) convolutional layer.
	}
	\label{fig:schematics}
\end{figure}

We consider a network with $L$ layers trained with a training set of input and output pairs $(\mathbf{x}, \mathbf{y})$. The number of nodes in the $l$th layer is $n_l$. The relationship between the output $x^l_i$ and the input $x^{l-1}_j$ of the $l$th layer are given as follows. The intermediate value $u_i^l$ is computed by
\begin{equation}
	u_i^{l} = \sum_{j=1}^{n_{l-1}} W_{ij}^l x_j^{l-1}
\end{equation}
for a fully connected layer, or
\begin{equation}
	u_i^{l} = \hbox{conv}(W^l_i, x^{l-1})
\end{equation}
for a convolutional layer. The output is activated by
\begin{equation}
	x_i^{l} = f^l(u_i^l;s^l_i)
\end{equation}
for $i=1,2,\cdots, n_l$, where $W_{ij}^l$ or $W^l_i$ are the weights. Instead of using a single activation function, we use  a set of activation functions in each layer. The activation function of the $l$th layer $f^l(\cdot ; s^l_i)$ takes the parameter $s^l_i$, which assigns a different activation function for each node. The selected activation functions are 
\begin{equation}
	f^l_i(u;s^l_i) = s^l_i f_0(u),
	\label{eq:activation} 
\end{equation}
which use the parameter set
\begin{equation}
	1 \geq s_1 \geq s_2 \geq \cdots, \geq s_{n_l} > 0,
	\label{eq:choice}
\end{equation}
where $f_0(\cdot)$ is an activation function such as a rectified linear unit (ReLU), hypertangent, or sigmoid function. The set of activation functions in each layer satisfy
\begin{equation}
	\frac{\partial f^l(u;s^l_i)}{\partial u} \geq 0
	\label{eq:cond1}
\end{equation}
for all $i$ and
\begin{equation}
	\frac{\partial f^l(u;s^l_i)}{\partial u} \geq \frac{\partial f^l(u;s^l_{i+1})}{\partial u}
	\label{eq:cond2}
\end{equation}
for $i=1,2,\cdots, n_l-1$ for all $u$. As a result, the nodes with smaller indices are assigned activation functions that have  increasingly steep slopes. Fig. \ref{fig:schematics} shows a schematic of the proposed network with a set of node-wise variant activation functions (ReLU activation functions in this case). 

A network can be trained to minimize a cost function $E$ that penalizes the difference between the network outputs and the desired outputs. When the network is trained using  backpropagation \cite{rumelhart1986learning, haykin1994neural}, the weight matrix $\mathbf{W}_l$, whose element is the weight $W_{ij}^l$ of the $l$th layer, is updated by
\begin{equation}
	\mathbf{W}_l \gets \mathbf{W}_l - \eta \bm{\delta}_l \mathbf{x}_{l-1}^\mathsf{T},
	\label{eq:update}
\end{equation}
where $\eta$ is the step size and $\mathbf{x}_{l-1}$ is a vector whose element is $x_i^{l-1}$. With the activation function given in \eqref{eq:activation}, the element of the sensitivity vector $\bm{\delta}_l$ is 
\begin{equation}
	\delta_i^L = s^L_i\frac{\partial E}{\partial x_i^L} \frac{\partial f^L_0 (u_i^L;s^l_i)}{\partial u_i^L} 
	\label{eq:sensitivity1}
\end{equation}
for the last layer and
\begin{equation}
	\delta_i^{l} = s^l_i (\sum_{k=1}^{n_{l+1}} \delta_k^{l+1} W_{kj}^{l+1}) \frac{f^l _0(u_i^{l};s^l_i)}{\partial u_i^l}
	\label{eq:sensitivity2}
\end{equation}
for the other layers. Because we adopt a set of activation functions whose first derivatives vary with the node indices. As a result, some nodes in a layer respond more sensitively than others when the weights are updated. In particular, the nodes with smaller node indices become increasingly more sensitive based on our choice of activation functions. We call such networks consisting of nodes with unequal and asymmetric sensitivities ``asymmetric networks.'' We call the parameter $s^l_i$, which affects the sensitivity of a node, the ``sensitivity variable.''

For activation functions given in the form of \eqref{eq:activation}, the weight matrix update in \eqref{eq:update} can be rewritten as follows:
\begin{equation}
	\mathbf{W}_l \gets \mathbf{W}_l - \eta (\mathbf{s}_l \circ \bm{\delta}_l^0) \mathbf{x}_{l-1}^\mathsf{T},
	\label{eq:update2}
\end{equation}
where $\mathbf{s}_l$ is a vector whose element is the parameter $s^l_i$, $\bm{\delta}^0_l$ is the sensitivity vector when a fixed activation function $f_0$ is used for all the nodes, and $\circ$ is the Hadamard product. The updated \eqref{eq:update2} in vector notation is 
\begin{equation}
	\mathrm{vec}(\mathbf{W}_l) \gets \mathrm{vec}(\mathbf{W}_l) - \eta \mathbf{S}_l \;\mathrm{vec}(\bm{\delta}_l^0 \mathbf{x}_{l-1}^\mathsf{T}),
	\label{eq:update3}
\end{equation}
where $\mathbf{S}_l$ is a diagonal matrix whose diagonal elements are repetitions of the vector $\mathbf{s}_l$. Note that \eqref{eq:update3} is the update for the diagonally scaled steepest descent algorithm resulting from the variable transforms
\begin{equation}
	\mathrm{vec}(\mathbf{W}_l) \gets \mathbf{S}_l^{\frac{1}{2}} \mathrm{vec}(\mathbf{W}_l)
	\label{eq:vartrans}
\end{equation}
for a quadratic penalty function \cite{bertsekas1999nonlinear}. The diagonal scaling is originally used to transform the variables to have equal dynamic ranges for faster convergence. We assume that the inputs of the network are already properly scaled, using normalization operation for example. By using the set of parameterized activation functions, we intentionally make the cost function steeper in some directions than others. The convergence in the steeper directions become much faster compared to that in the other, less steep directions. In particular, we make the cost function steeper in directions associated with the weights of sensitive nodes than in directions associated with weights of insensitive nodes. As a result, asymmetric networks learn weight of sensitive nodes much faster than that those of insensitive nodes.

\subsection{Analysis of Asymmetric Networks}
\label{sec:analysis}

For analysis, consider an asymmetric network with one hidden layer given by
\begin{equation}
	\hat{\mathbf{y}} = \mathbf{W}_2 \mathbf{D}f_0(\mathbf{W}_1\mathbf{x}) , 
	\label{eq:linearmodel}
\end{equation}
where $\mathbf{x} \in \mathbb{R}^n$ and $\hat{\mathbf{y}} \in \mathbb{R}^m$ are the input and output of the network, respectively. The matrices $\mathbf{W}_1 \in \mathbb{R}^{p\times n}$ and $\mathbf{W}_2 \in \mathbb{R}^{m\times p}$ represent the operations of the first and second layers, respectively. The rows of $\mathbf{W}_2$ and $\mathbf{W}_1$ contain the weights for the output nodes of the layers. The function $f_0$ and the diagonal matrix $\mathbf{D}\in\mathbb{R}^{p\times p}$ represent the operation of the activation function in \eqref{eq:activation}. The diagonal elements of the matrix $\mathbf{D}$ are the sensitivity parameters, $s_i$'s.

Let $\mathbf{X}$ and $\mathbf{Y}$ be matrices whose columns are input and output vectors in a training set, respectively. The covariance matrices of input and output are $\Sigma_\mathbf{xx}$ and $\Sigma_\mathbf{yy}$, respectively, and the cross-covariance matrix is $\Sigma_\mathbf{yx}$. The network is trained with the cost function
\begin{equation}
	E = \| \mathbf{Y} - \mathbf{W}_2 \mathbf{D}f_0(\mathbf{W}_1\mathbf{X}) \|^2.
	\label{eq:analysiscost}
\end{equation}
The cost function is minimized when the operation of the network is equivalent to the slope matrix of the least square regression of $\mathbf{Y}$ on $\mathbf{X}$ \cite{baldi1989neural}.
Let $\mathbf{u}_1, \mathbf{u}_2, \cdots, \mathbf{u}_r$ be the eigenvectors of $\Sigma$ with corresponding eigenvalues $\lambda_1 \geq \lambda_2 \geq \cdots \geq \lambda_r$. We consider a case $p>r$, where the number of hidden nodes is larger than the rank of $\Sigma$. Let
\begin{equation}
	\mathbf{U} = \left[  \mathbf{u}_1, \mathbf{u}_2, \cdots, \mathbf{u}_r, \mathbf{0},\cdots,\mathbf{0}\right].
\end{equation}
When the minimum of the cost function is achieved, the operation of the network is equivalent to
\begin{equation}
	\mathbf{W}_2 \mathbf{D}f_0(\mathbf{W}_1\mathbf{x}) \approx \mathbf{U}\mathbf{U}^\mathsf{T}   
\end{equation}
The sensitivity parameter $s_i$ is close to one for sensitive nodes, and close to zero for insensitive nodes. 
\begin{eqnarray}
	\sum_{i\in {\mathcal S}}  s_i \mathbf{w}^2_i  f_0((\mathbf{w}^1_i)^\mathsf{T} \mathbf{x})& \approx & \sum_{i=1}^r \mathbf{u}_i \mathbf{u}_i^\mathsf{T}  \\
	 \sum_{i\in {\mathcal I}} s_i \mathbf{w}^2_i  f_0((\mathbf{w}^1_i)^\mathsf{T} \mathbf{x})	& \approx &  \mathbf{0} 
\end{eqnarray}
where $\mathbf{w}_i^2$ and $\mathbf{w}_i^1$ are the $i$th columns of $\mathbf{W}_2$ and $\mathbf{W}_1$, respectively, and $\mathcal{S}$ and $\mathcal{I}$ are the index sets of sensitive nodes and insensitive nodes. The sensitive nodes are mostly responsible for the operation of the network in approximating the projection onto the eigenvectors $\mathbf{u}_i$'s, and the insensitive nodes does not contribute much to the approximation. 

The asymmetric network leans the weight of sensitive nodes faster than insensitive nodes. Consider the following extreme case. The sensitivity variables are chosen to be
\begin{equation}
	\mathbf{s}_1^\mathsf{T} = [1,\epsilon,\cdots, \epsilon]^\mathsf{T},
\end{equation}
where $\epsilon$ is close to 0. Then the operation of the network approximates
 \begin{equation}
	 s_1 \mathbf{w}^2_1  f_0((\mathbf{w}^1_1)^\mathsf{T} \mathbf{x}) \approx \sum_{i=1}^r \mathbf{u}_i \mathbf{u}_i^\mathsf{T},\end{equation}
and the weight of the first node approximates the eigenvector $\mathbf{u}_1$ corresponding to the largest eigenvalue. After the convergence, the sensitivity variables are changed to
\begin{equation}
	\mathbf{s}_1^\mathsf{T} = [1,1, \epsilon,\cdots, \epsilon]^\mathsf{T}.
\end{equation}
The operation of the network approximates
 \begin{equation}
	 s_2 \mathbf{w}^2_2  f_0((\mathbf{w}^1_2)^\mathsf{T} \mathbf{x}) \approx \sum_{i=2}^r \mathbf{u}_i \mathbf{u}_i^\mathsf{T},\end{equation}
and the weight of the second node approximates the eigenvector $\mathbf{u}_2$ corresponding to the second largest eigenvalue. This process can continue to find the weights of the $r$ nodes. 

When the sensitivity variables are chosen as \eqref{eq:choice}. The cost function is steeper in directions of weights of nodes with smaller indices.  The weights of the nodes with smaller indices can be expected to converge faster than those with larger indices. Based on the above observation in the extreme case---where the node weights are converged individually---we can assume that the features learned by the network are sorted in order of their importance: from the first hidden node to the last hidden node.

Deep networks with nonlinear activation functions trained by the backpropagation are not quite as straightforward to analyze as is the shallow network analyzed in this section. There will be some difficulties associated with the nonlinearity of the activation functions, random initialization, and the herd effect of the backpropagation. Nevertheless, the cost functions are minimized when the weight matrices are related to the least squares regression. For asymmetric deep networks, the cost functions are minimized when eigenvectors that correspond to larger eigenvalues are assigned as weight vectors of nodes with larger sensitivity parameters.

\section{Pruning Deep Asymmetric Networks for Efficient Design}
\label{sec:pruning}

\subsection{Pruning Deep Asymmetric Networks}
\label{sec:pruningalgorithm}

Deep networks are typically computationally expensive and have high memory requirements. In many cases, the complicated models of deep networks must be compressed to be deployed on devices with restricted computational capabilities. Previous studies show designing compact and efficient networks is possible by pruning unimportant nodes. However, determining which node is important  from a trained network is not obvious, because the nodes in a network have a tendency to share workloads among themselves \cite{baldi1989neural}.

In the proposed asymmetric networks, nodes in a layer are sorted in order of their importance. Thus, given an asymmetric network with $p$ hidden nodes, when asked to remove one hidden node, we can know exactly which node to discard. The last hidden node, which corresponds to the least important feature, can be removed from the network with the least accuracy loss. Hence, a simple and effective pruning strategy for designing efficient deep asymmetric network can be derived.

First, the accuracy of a trained asymmetric network is measured using a validation set. Then, the network is pruned by removing the nodes individually to ensure that the network accuracy does not fall below a certain level. The pruning process can starts from the last hidden layer and continues up to the first layer. The nodes in each layer are discarded one-by-one from last to first. The accuracy of the network is measured after removing each node using the validating set. The pruning of a current layer stops just before the accuracy drops below a predefined threshold; then, the pruning continues with the previous layer, again discarding nodes from last to first. Our empirical data indicates that pruning works better when it begins with a layer that has the largest number of nodes and continues with layers with smaller and smaller number of nodes. Thus, we prepare a set of layer indices $\mathcal{L}$ sorted by the number of nodes, and prune a network following the layer index in the set. The pseudocode of this pruning strategy is listed in Algorithm \ref{alg:pruning}. Note that the algorithm does not require any measure, such as $l_2$ or $l_1$, to evaluate the importance of a node. The importance of nodes is already given by the node indices. The algorithm simply prune the node sequentially while the target accuracy is met.

\begin{algorithm}[t!]
\caption{Pruning}
\label{alg:pruning}
\begin{boxedminipage}{\linewidth}
	\begin{algorithmic}[1]
	\STATE Measure \texttt{accuracy} and set \texttt{target}.
	\FOR{$\zeta = 1$ to $L-1$}
	\STATE $l = \mathcal{L}(\zeta)$
	\FOR {$i=n_l$ to $1$}
	\STATE Measure \texttt{accuracy}
	\IF{$\hbox{\texttt{accuracy}} \leq \hbox{\texttt{target}}$} 
	\STATE \textbf{break}
	\ELSE
	\STATE Remove node $x^l_i$ from network.
	\ENDIF
	\ENDFOR
	\ENDFOR
	\end{algorithmic}
\end{boxedminipage}
\end{algorithm}

Once nonessential nodes are pruned by the Algorithm \ref{alg:pruning}, the network is retrained using the training set. Because all the important features are retained by the pruned network, the retraining can compensate performance loss due to the pruned nodes. As a result, a network with the same performance but with far less nodes can be designed by the pruning and retraining procedure.

\subsection{Review of Pruning Methods}
\label{sec:pruningreview}

A survey of efficient deep network design methods can be found in \cite{cheng2017survey, reed1993pruning}. In many pruning approaches, the importance of nodes in a trained network are evaluated with a certain measure, and the nodes showing smaller measures are removed from the network. In \cite{lecun1990optimal, hassibi1993second}, the sensitivity of a cost function to small changes in weights of a network is measured based on the Hessian of the cost function. The weights are sorted by sensitivity; then, the nonessential small-sensitivity weights are pruned from the network. The sensitivity to small changes in node outputs \cite{mozer1989skeletonization} and the sensitivity to the removal of weights \cite{karmin1990simple} have also been considered for pruning. A regularization term that penalizes large weights can be added to a cost function during network training to favor sparse weights. After training, the weights are sorted by some measure, and small weights are removed from the trained network. The $l_2$, $l_1$, and $l_0$ norms, among other, have been used as regularization terms to cause  weights to be sparse to enable the pruning of small weights from networks \cite{han2015learning, ishikawa1996structural, collins2014memory, li2016pruning, zhou2016less}. The weight decay term was used as a regularization factor to favor small weights in \cite{hanson1989comparing}. The energy term was  used as a regularization term to penalize nodes with small outputs in \cite{chauvin1989back}. After the nonessential nodes with small sensitivity or small weights have been pruned, the networks are usually retrained to regain their performance. The pruning and retraining procedures can be applied iteratively. The performance of pruned networks depends on how effectively the sensitivity or the regularization identify the nonessential weights or nodes. However, because the nodes in deep asymmetric networks are already sorted in order of their importance, the nonessential nodes can be removed from the network without requiring further inspection of their importance. In \cite{lebedev2016fast}, sparsity patterns were introduced that  reduced the computations of convolutional networks by zeroing out filter coefficients. In \cite{han2015deep}, the filter coefficients were quantized and compressed to reduce the memory requirements.

\section{Experiments}
\label{sec:ex}

The proposed asymmetric network is analysed with an autoencoder using Gaussian, the MNIST dataset \cite{lecun1998gradient}, and with a deep CNN using the CIFAR-10 dataset \cite{krizhevsky2009learning} in Section \ref{sec:exgauss}, \ref{sec:exan} and \ref{sec:exdan}, respectively. The performance of a pruned asymmetric network is compared to other pruning approaches with LeNet \cite{krizhevsky2012imagenet} using the MNIST dataset in Section \ref{sec:lenet}. Asymmetric networks with weights transferred from trained networks are analyized with VGG  \cite{simonyan2014very} and ResNet \cite{he2016deep} using the CK+ facial expression recognition dataset \cite{lucey2010extended} in Section \ref{sec:ck+}, and with a deep CNN using the NTU RGB+D action recognition dataset in Section \ref{sec:NTU} \cite{du2015skeleton}. The performance of pruned asymmetric networks, with the weight transfer, is compared to other pruning approaches with VGG and ResNet using the CIFAR-10 dataset in Section \ref{sec:CIFAR}, and with ResNet using the ImageNet dataset \cite{deng2009imagenet} in Section \ref{sec:ImageNet}.

\subsection{Simple Asymmetric Network with Gaussian Data}
\label{sec:exgauss}

In this section, a simple autoregression experiment is set up with a three dimensional data following the Gaussian distribution. The input $\mathbf{x}=[x_1,x_2,x_3]^\mathsf{T}$ follows the zero mean Gaussian distribution. The elements $x_1$ and $x_2$ are correlated with 0.9 correlation coefficient. The element $x_3$ is independent. Standard deviations of the three elements are set to 1.0, 1.0, and 0.01. An asymmetric network with one hidden layer with four hidden nodes is prepared. The weights of the first and second layers are tied together. The activation function $f_0$ is set to a linear function. A symmetric network with the same architecture is also prepared. The networks are trained with 2048 data by solving \eqref{eq:analysiscost} for autoregression. The Matlab optimization toolbox is used to find the solutions. The optimization function is initialized randomly.

\begin{figure}[t!]
	\centering
	\begin{minipage}{\linewidth}
		\centering
		{\includegraphics[width=0.33\linewidth]{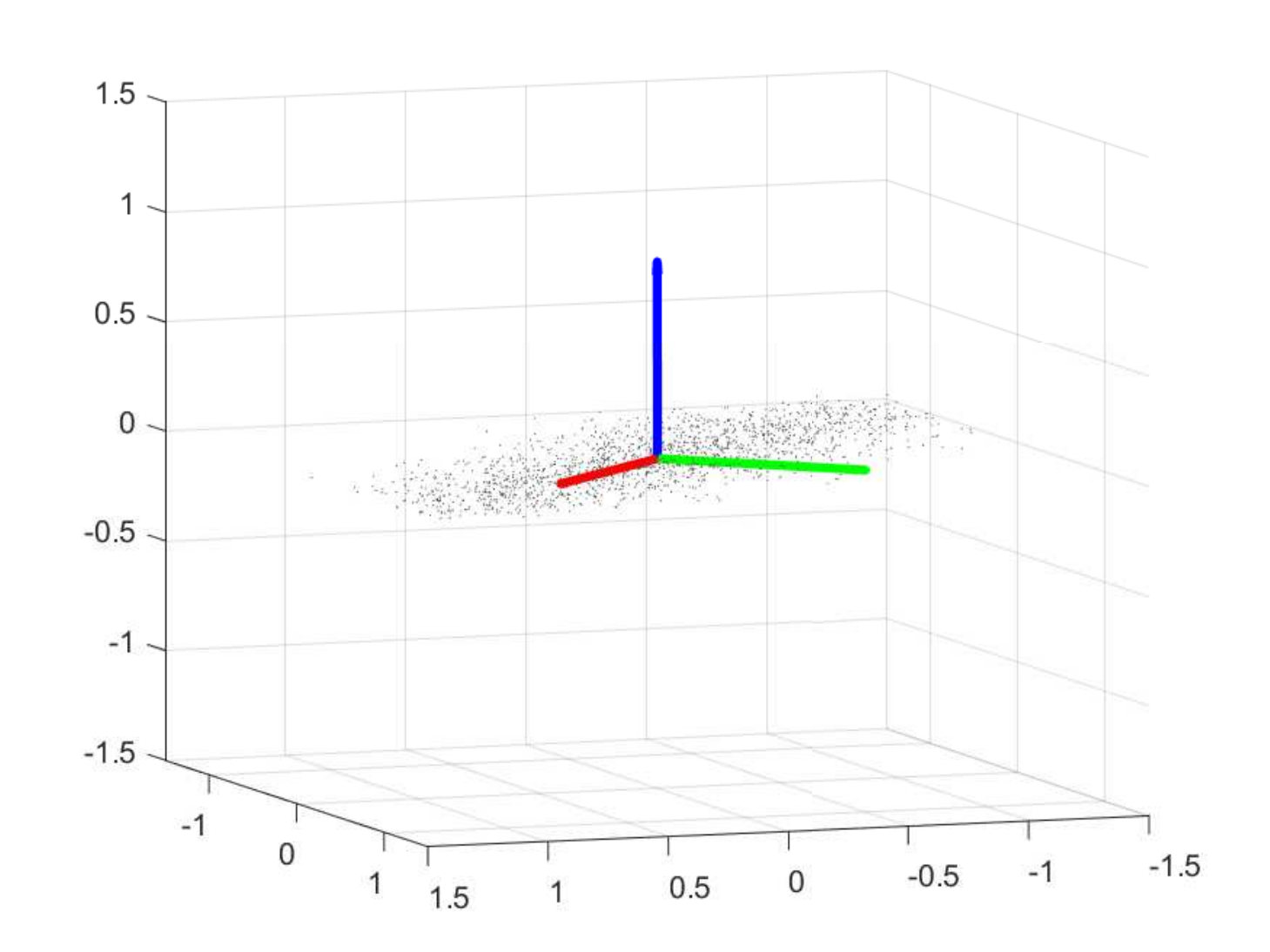}}%
		{\includegraphics[width=0.33\linewidth]{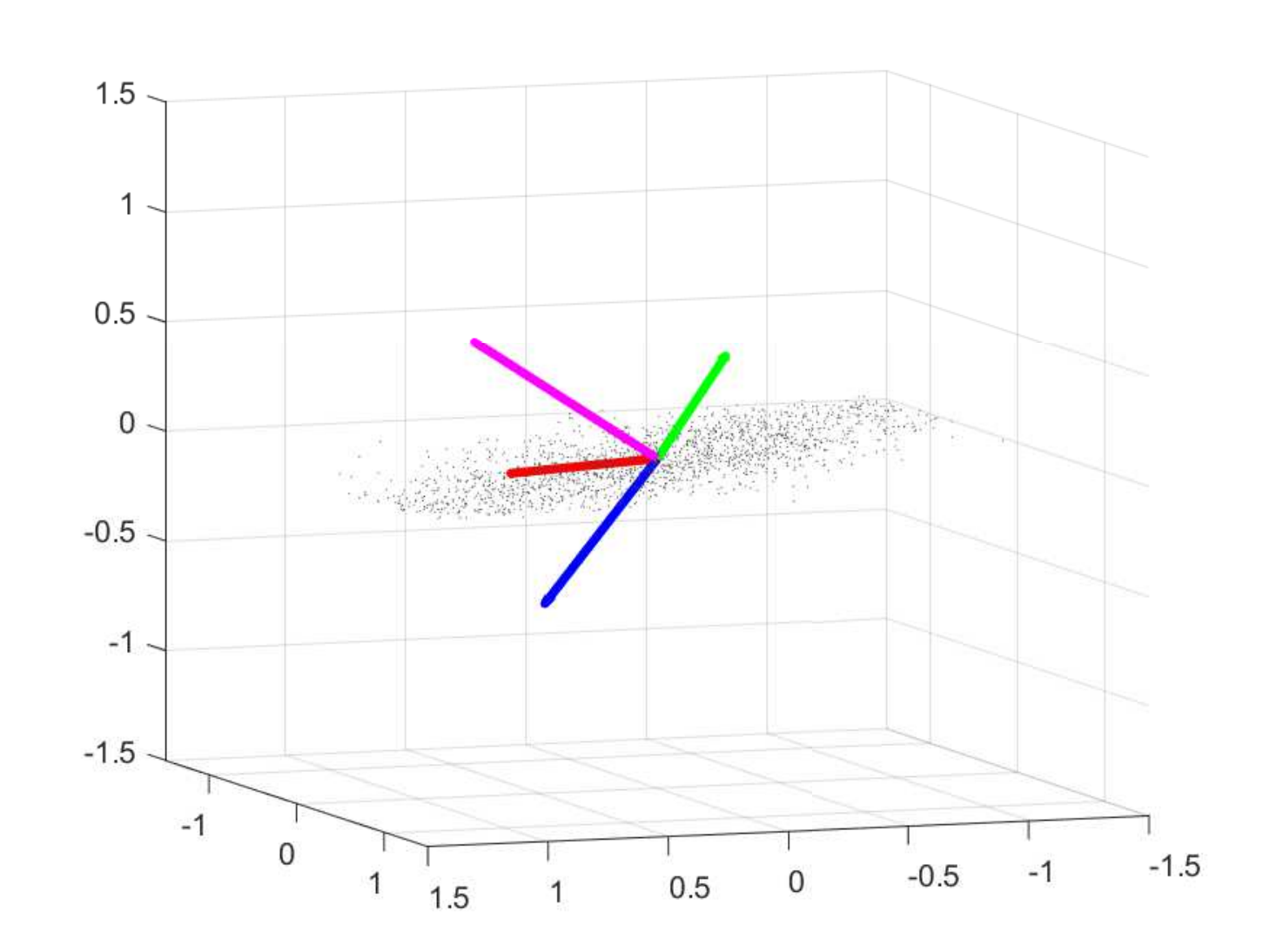}}%
		{\includegraphics[width=0.33\linewidth]{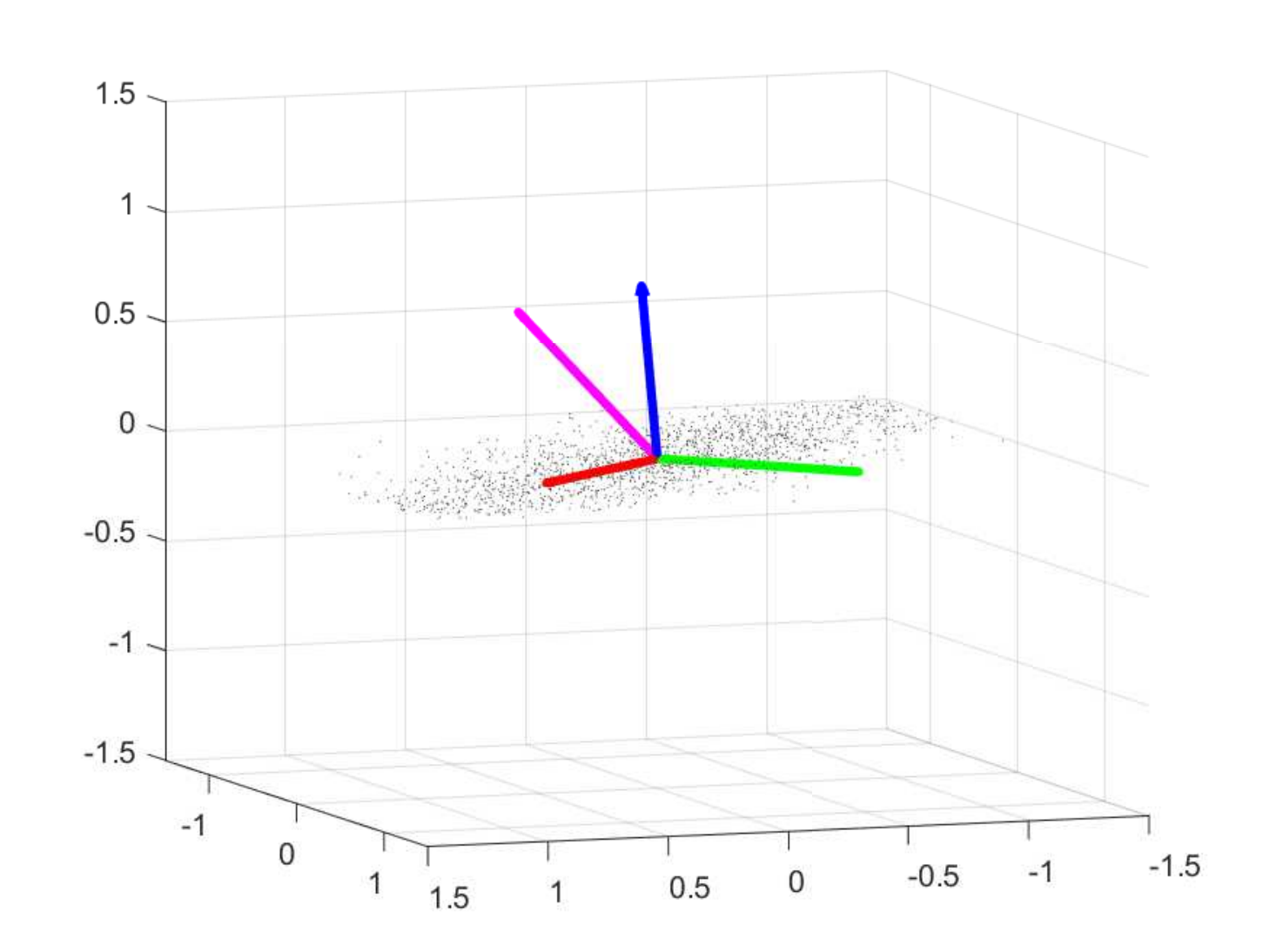}}%
		
		{\footnotesize (a)}
	\end{minipage}

	\begin{minipage}{\linewidth}
		\centering
		{\includegraphics[width=0.33\linewidth]{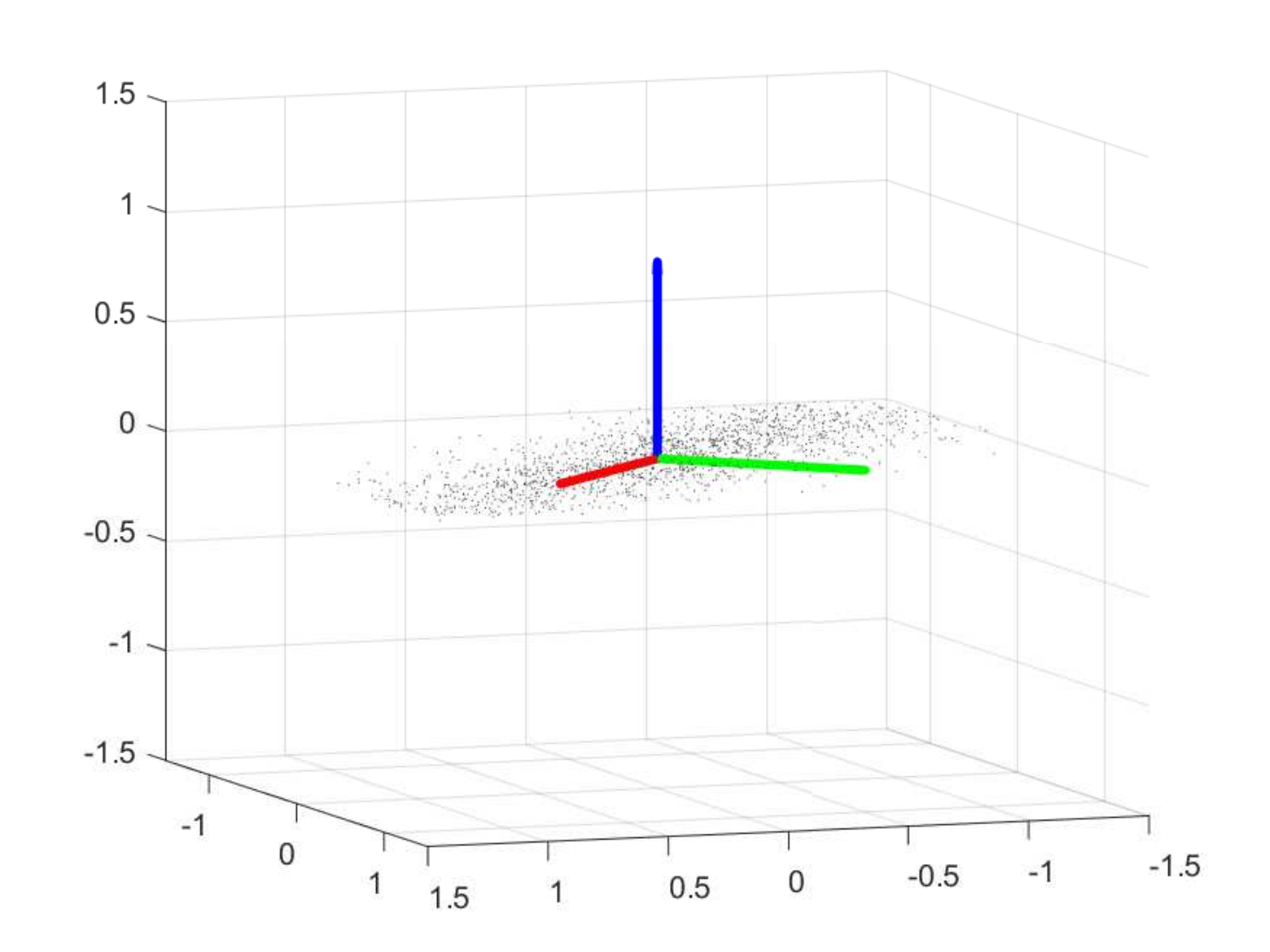}}%
		{\includegraphics[width=0.33\linewidth]{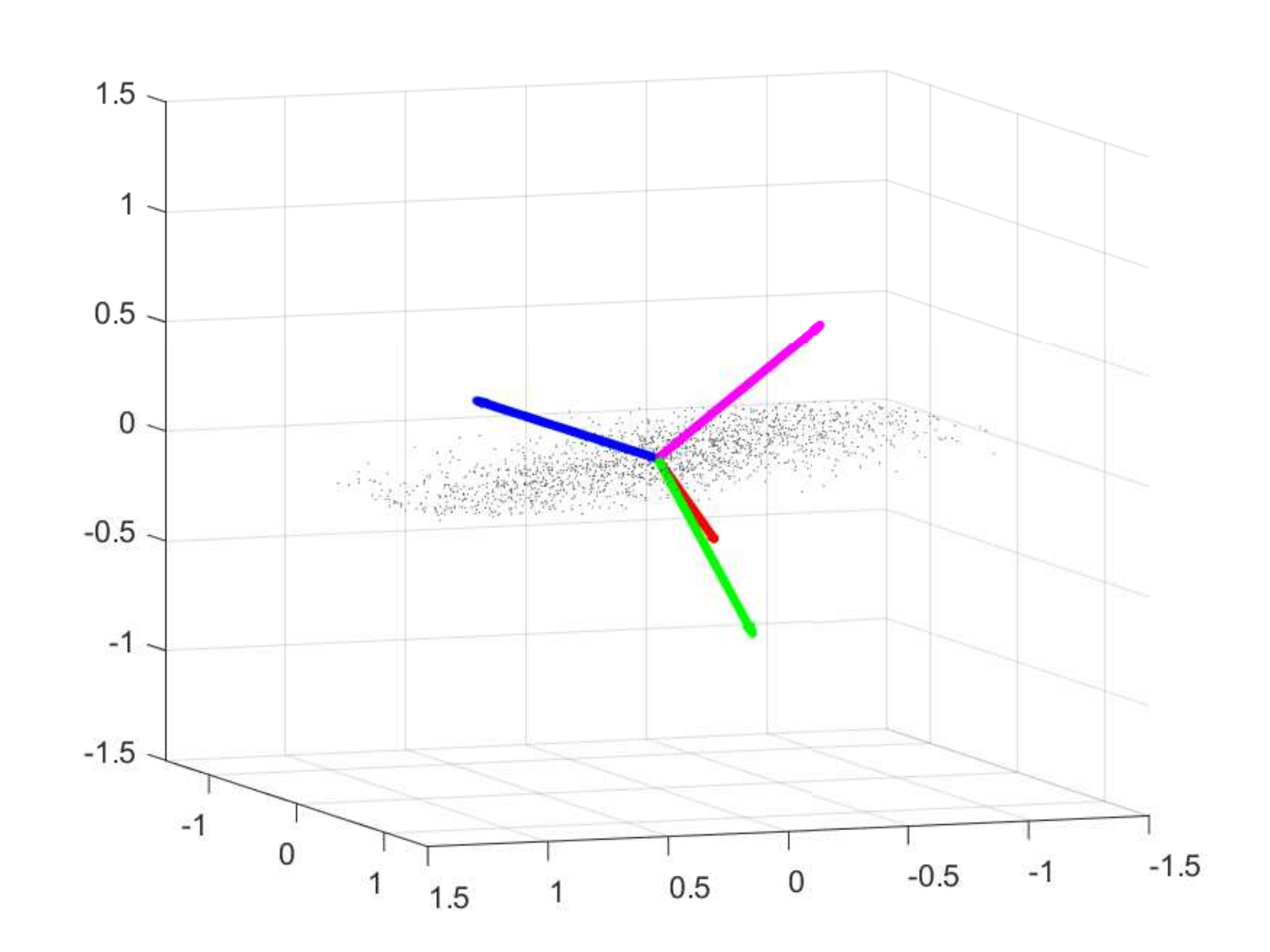}}%
		{\includegraphics[width=0.33\linewidth]{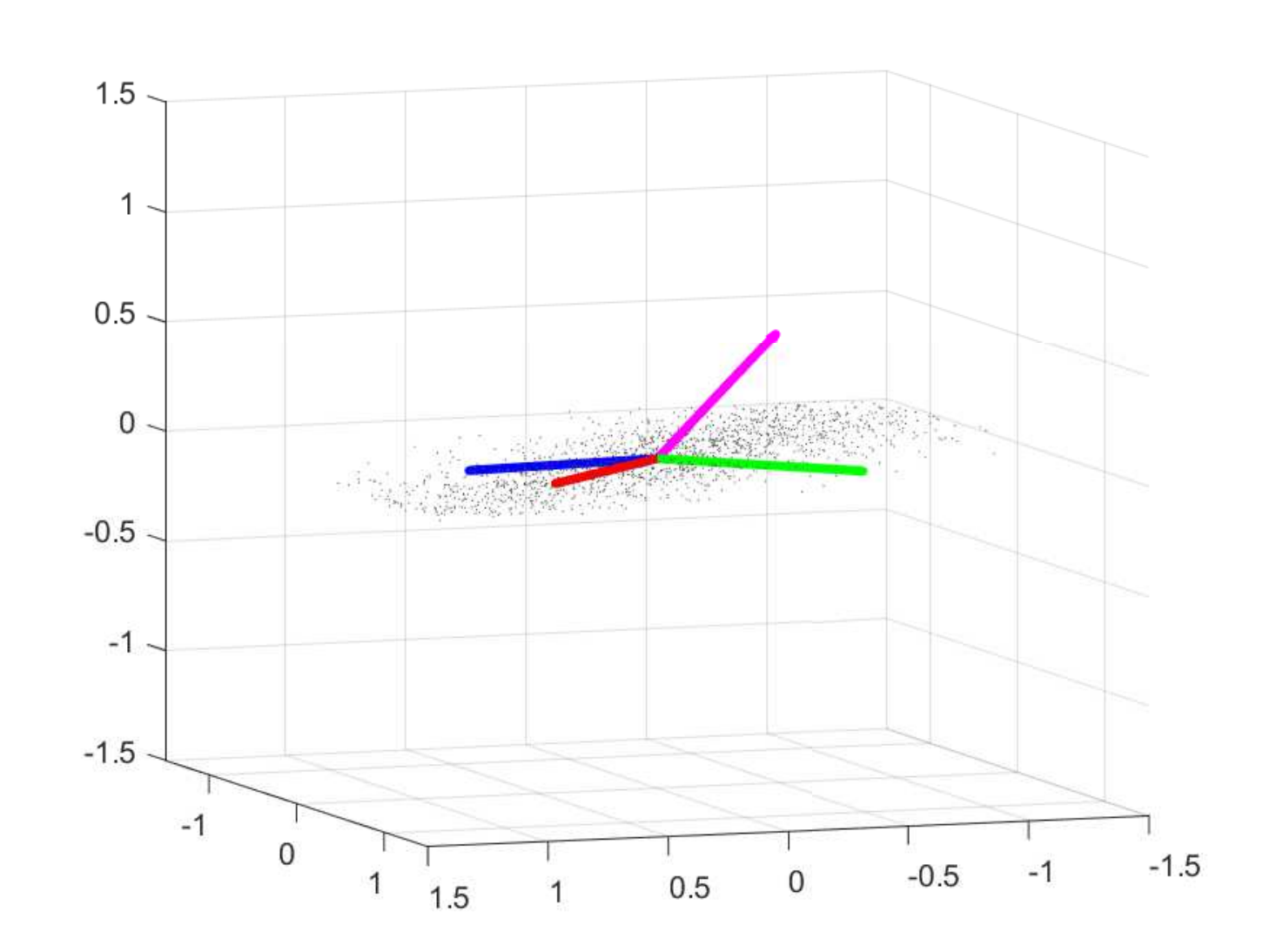}}%
		
		{\footnotesize (b)}
	\end{minipage}
	\caption{
		Weight vectors of hidden nodes for Gaussian dataset:
		(a) for highly heterogeneous sensitivity variables;
		(b) for moderately heterogeneous sensitivity variables.
		Left: eigenvectors;
		center: weight vectors of symmetric network; and
		right weight vectors of asymmetric network.
		The 1st, 2nd, 3rd, and 4th nodes are labeled with red, green, blue, and magenta, respectively.
	}
	\label{fig:exgauss}
\end{figure}

Fig. \ref{fig:exgauss} shows the normalized weight vectors of the networks. For comparison, the eigenvectors of the covariance matrix $\Sigma_\mathbf{xx}$ are also shown. Fig. \ref{fig:exgauss}(a) shows the case where the sensitivity variables are chosen to be highly heterogeneous with $\mathbf{s}_1^\mathsf{T} = [1,(1/2)^8,(1/4)^8,(1/8)^8]^\mathsf{T}$. For the asymmetric network, the weight vectors of the first and second nodes coincide with the eigenvectors corresponding to the largest and second largest eigenvalues. The weight vectors of the third and fourth nodes span the space orthogonal to the column space of the first two eigenvectors. The asymmetric network learns to represent the data using the two sensitive nodes. Among the two sensitive nodes, the first node is related to the more principal component. In comparison, the weight vectors of the symmetric network are not aligned to the eigenvectors. Fig. \ref{fig:exgauss}(b) shows the case where the sensitivity variables are chosen to be moderately heterogeneous with $\mathbf{s}_1^\mathsf{T} = [1,(1/2)^6,(1/4)^6,(1/8)^6]^\mathsf{T}$. In this case, the weight vectors of the first three nodes span the columns space of the first two eigenvectors. Nevertheless, the weight vectors of the first and second nodes coincide with the eigenvectors corresponding to the largest and second largest eigenvalues. Because of the random initialization, the weight vectors vary every time the solution is found. However, the trend is consistent for the highly heterogeneous case, and mostly consistent for the moderately heterogeneous case. We will discuss the consistency in terms of repeatability and reproducability using a deep network with a real dataset in Section \ref{sec:exan}.

\subsection{Asymmetric Networks with MNIST Dataset}
\label{sec:exan}

\begin{figure}[t!]
	\centering		
	\begin{minipage}{0.33\linewidth}		
		\centering
		{\includegraphics[width=\linewidth]{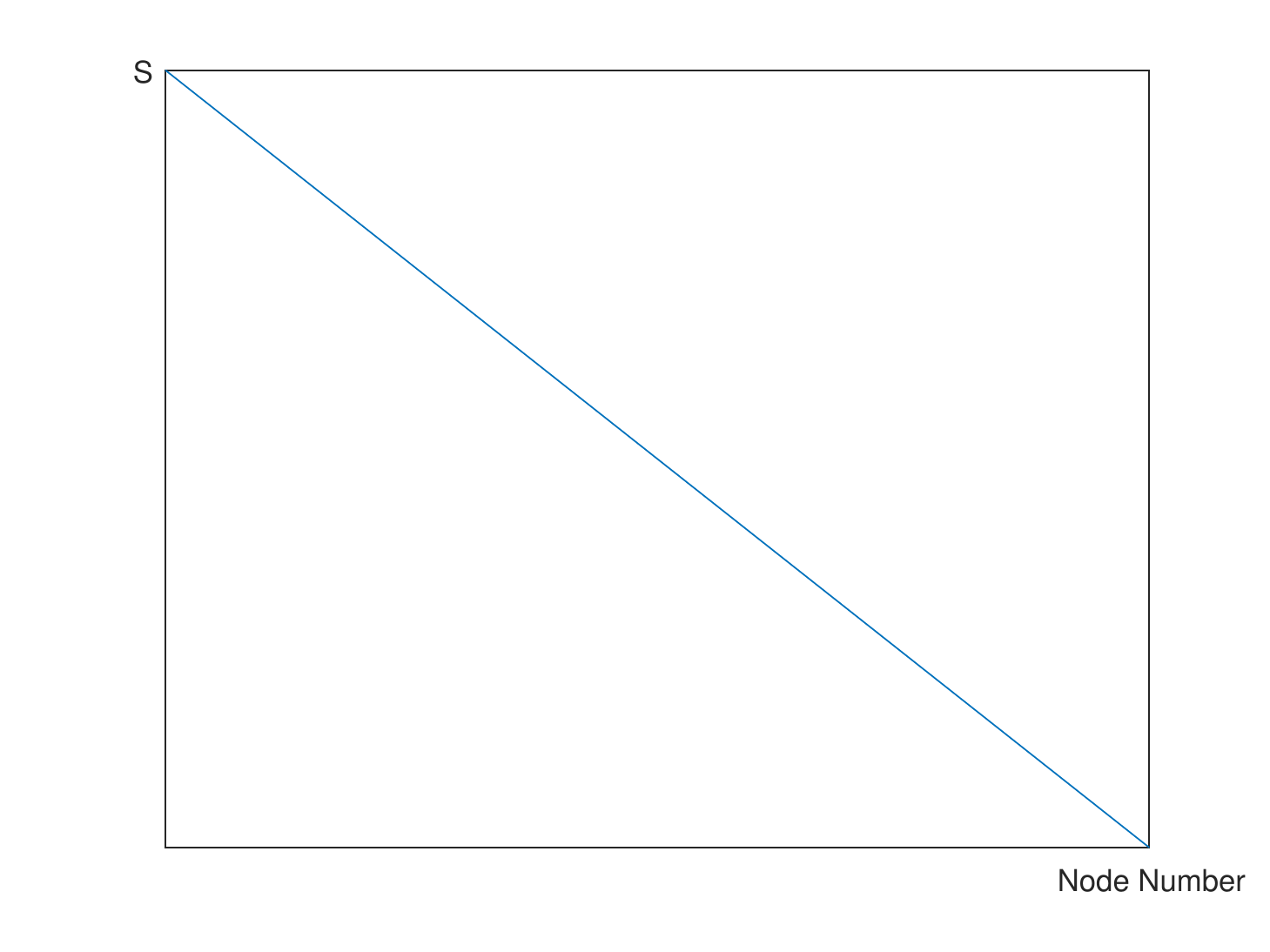}}%
		
		{\footnotesize (a)}
	\end{minipage}%
	\begin{minipage}{0.33\linewidth}		
		\centering
		{\includegraphics[width=\linewidth]{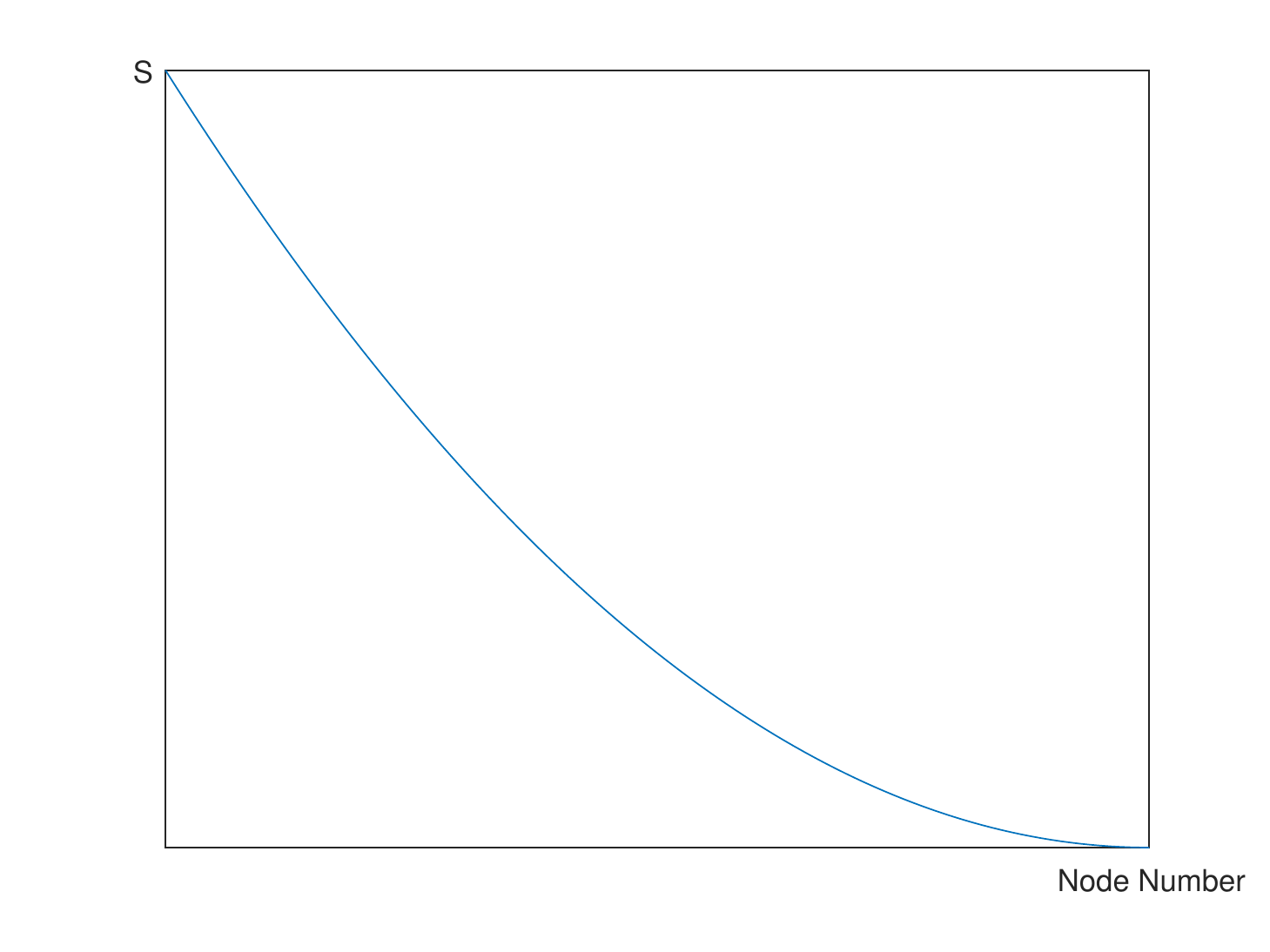}}%
		
		{\footnotesize (b)}
	\end{minipage}%
	\begin{minipage}{0.33\linewidth}		
		\centering
		{\includegraphics[width=\linewidth]{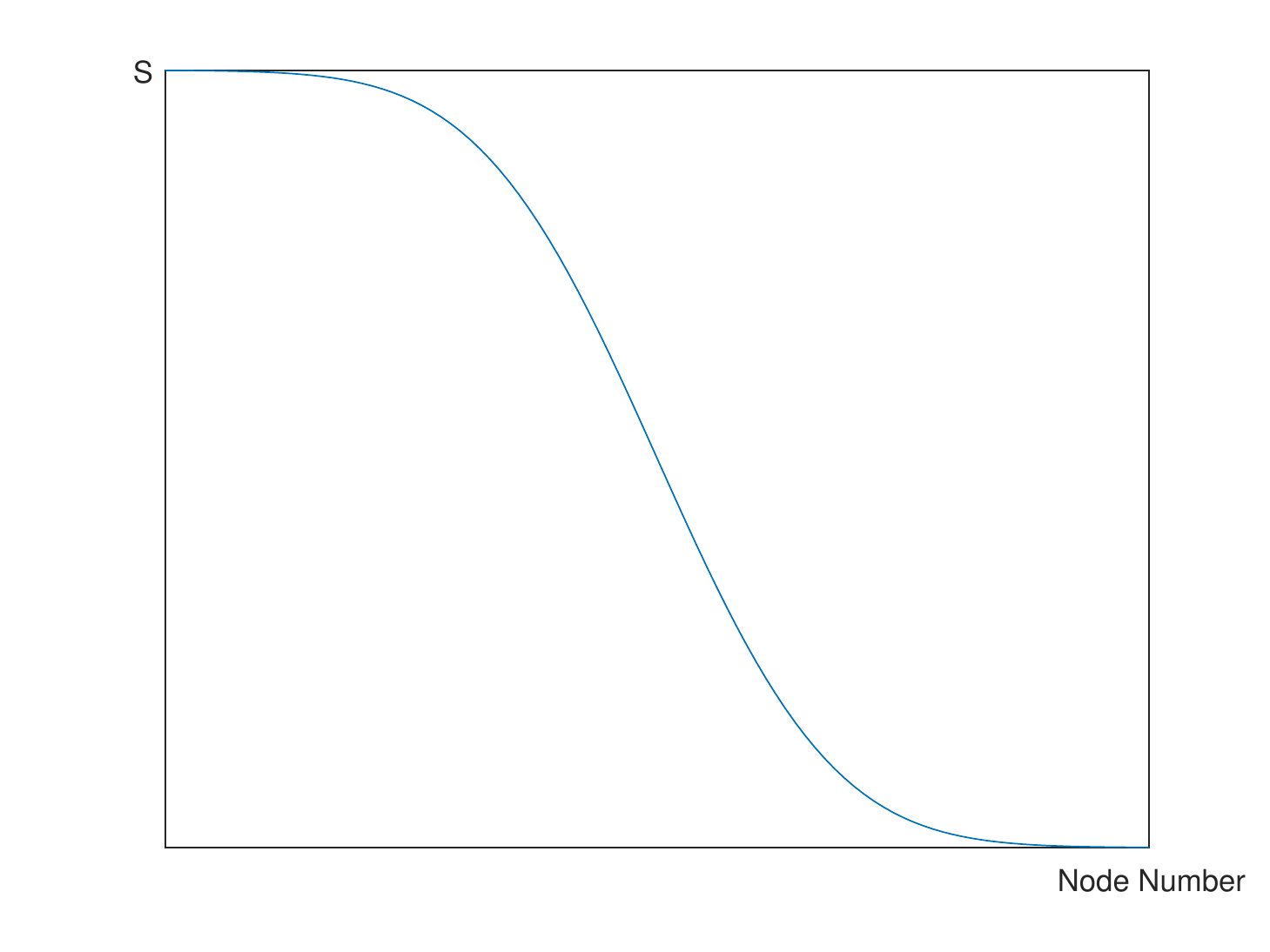}}%
		
		{\footnotesize (c)}
	\end{minipage}%
		
	\caption{The sensitivity parameter $s_i$ used for a set of node-wise variant ReLU activation functions. 
	}
	\label{fig:parameters}
\end{figure}

The use of a set of node-wise variant activation functions in an asymmetric network is validated with a simple fully connected network with one hidden layer using the MNIST dataset \cite{lecun1998gradient}. A network with $n$ hidden nodes is prepared in an auto-associative setting to reconstruct the input. The number of hidden node $n$ is set to 256. A set of activation functions with the ReLU function as $f_0$ is used. The sensitivity parameters $s_i$ are set as shown in Fig. \ref{fig:parameters}(b) (this parameter choice is discussed in Section \ref{sec:exdan}). The network is implemented using the Keras Python deep learning library. A baseline symmetric network, with the same number of hidden nodes and the ReLU function as activation functions for all the hidden nodes, is prepared for comparison. 

\begin{figure}[t!]
	\centering		
	\begin{minipage}{0.5\linewidth}		
		\centering
		{\includegraphics[trim = 50 40 55 10, clip, width=\linewidth]{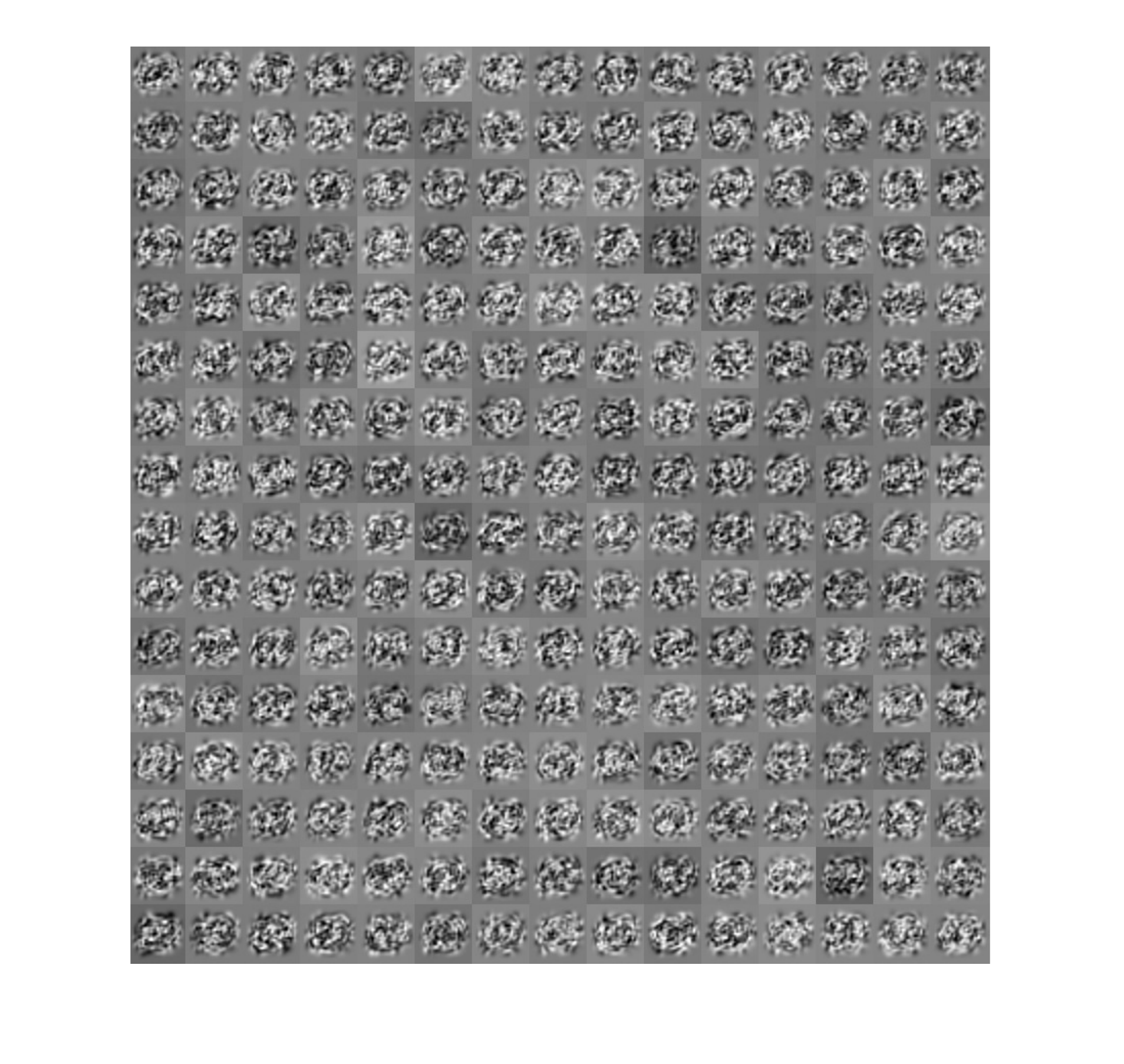}}%
		
		{\footnotesize (a)}
	\end{minipage}%
	\begin{minipage}{0.5\linewidth}		
		\centering
		{\includegraphics[trim = 50 40 55 10, clip, width=\linewidth]{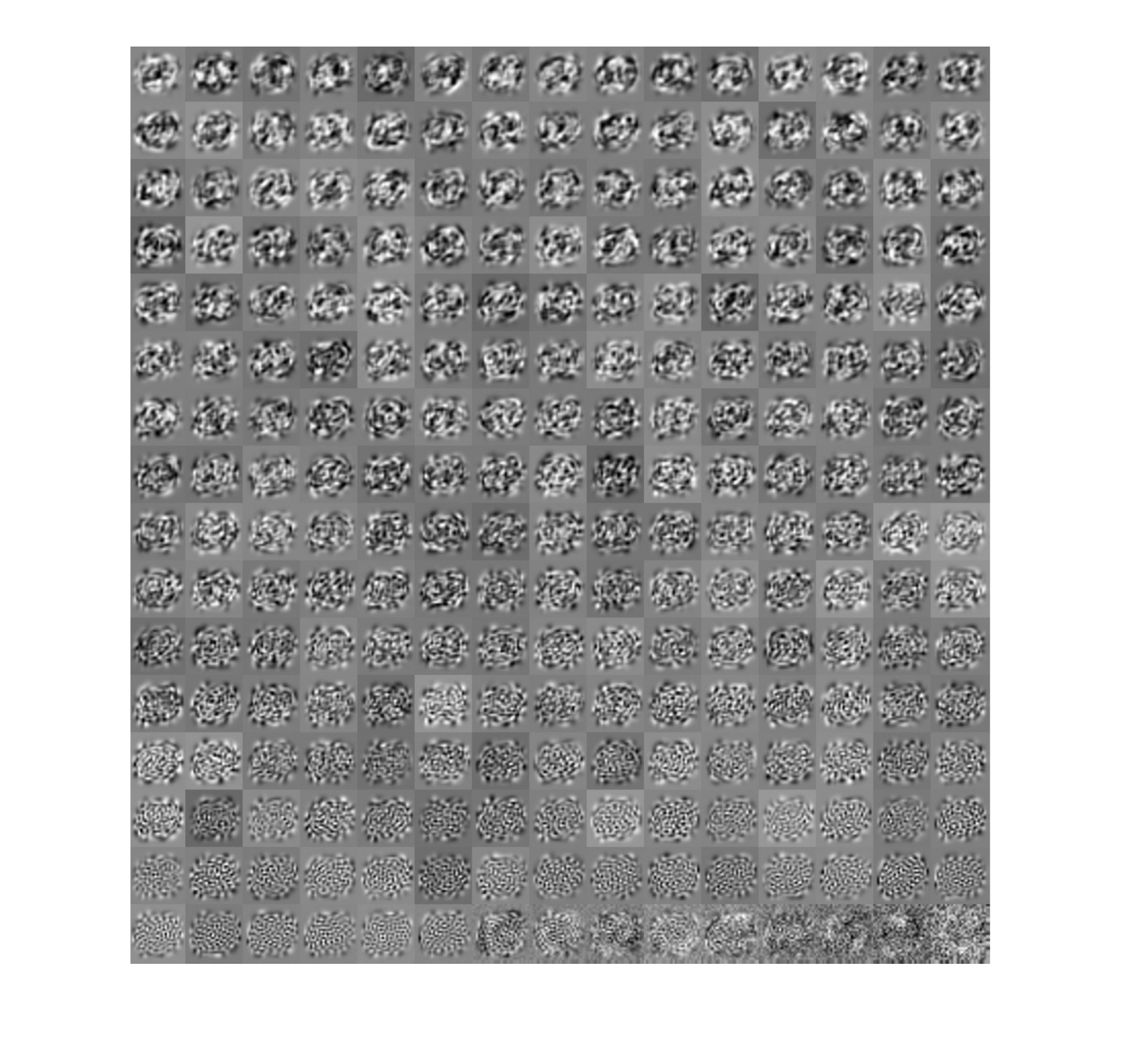}}%
		
		{\footnotesize (b)}
	\end{minipage}%
		
	\caption{Features learned by the networks with MNIST dataset,
	(a) baseline symmetric network, and
	(b) asymmetric network.
	}
	\label{fig:MNISTfeatures}
\end{figure}

Fig. \ref{fig:MNISTfeatures} shows the features learned by the baseline and asymmetric networks. The two networks are initialized to the same weights and trained using the same training set. For the baseline symmetric network, features of all the nodes that appeared in the early iterations of the training becomes more clear as training progresses. In contrast, for the asymmetric network, features of the nodes with high indices that appeared at the early iterations looses their structures and becomes more high frequency noise-like. Features with clearer structures are concentrated to the nodes with lower indices.

\begin{figure}[t!]
	\centering		
	\begin{minipage}{0.5\linewidth}		
		\centering
		{\includegraphics[width=\linewidth]{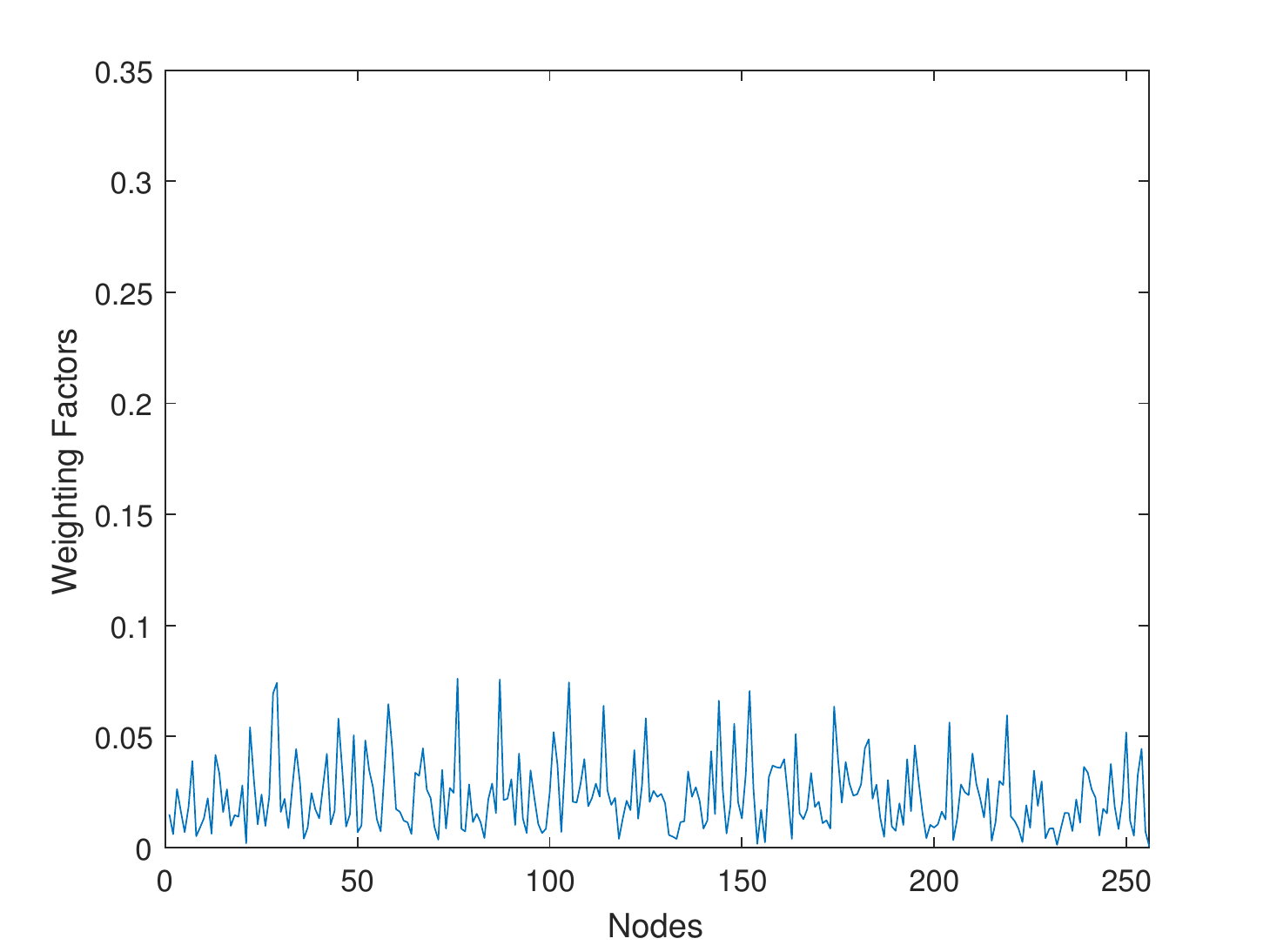}}%
		
		{\footnotesize (a)}
	\end{minipage}%
	\begin{minipage}{0.5\linewidth}		
		\centering
		{\includegraphics[width=\linewidth]{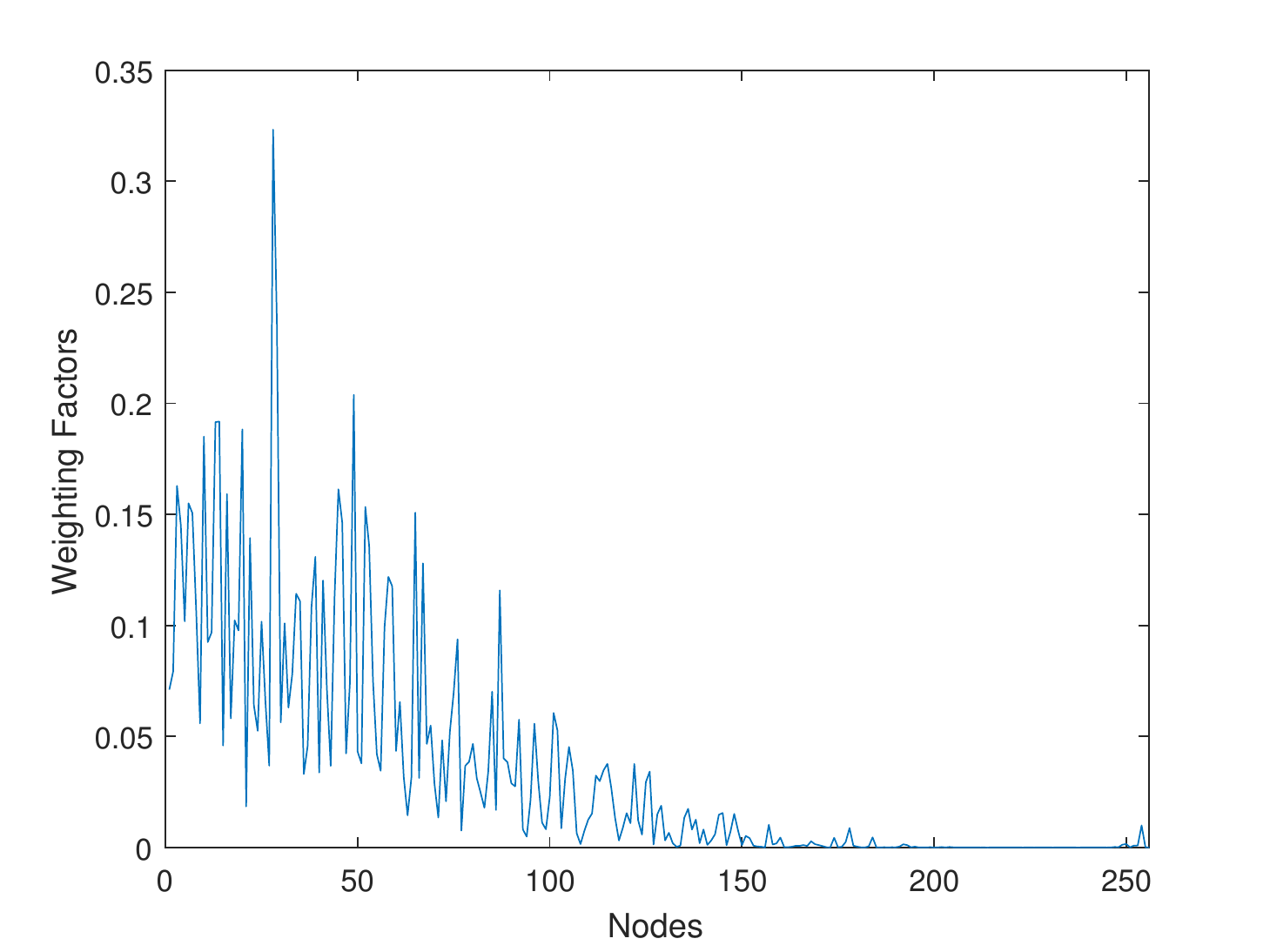}}%
		
		{\footnotesize (b)}
	\end{minipage}%
		
	\caption{Approximation of data using the features learned by the networks. 
	The average weighting factors to represent data as linear combinations of features are shown. 
	(a) baseline symmetric network, and
	(b) asymmetric network.
	}
	\label{fig:MNISTapprox}
\end{figure}

The weights of the networks are related to the eigenvectors of the covariance matrix $\Sigma_\mathbf{xx}$ of the training data. We represent images in the training set as linear combinations of the weights. The weights of the second layer, $\mathbf{w}^2_i$, are normalized to $\tilde{\mathbf{w}}^2_i$ such that $\|\tilde{\mathbf{w}}^2_i\|_2=1$. The weighting factors for the linear combination, $\mathbf{z}=[z_1,z_2, \cdots, z_n]^\mathsf{T}$, to represent an image $\mathbf{y}$ is found by
\begin{equation}
	\underset{\mathbf{z}}{\hbox{minimize}} \; \| \mathbf{y} - \left[\tilde{\mathbf{w}}^2_1,\tilde{\mathbf{w}}^2_2, \cdots, \tilde{\mathbf{w}}^2_n\right] \mathbf{z}\|_2 + \mu \|\mathbf{z}\|_1 
\end{equation}
The $\ell_1$ norm on $\mathbf{z}$ is used as a regularizer with a small value for the regularization parameter $\mu$, so that the image is represented as a sparse linear combination of non-orthogonal weights $\tilde{\mathbf{w}}^2_i$. Fig. \ref{fig:MNISTapprox} shows the weighting factor $\mathbf{z}$ for the symmetric and asymmetric networks. The average weighting factors to represents 30 images in the training set are shown. For the baseline network, images are represented using all the weights. The weighting factor $\mathbf{z}$ have significant values for all the weights. The features the baseline network learned are all of the same importance. In contrast, for the asymmetric network, images are represented using only the weights with the small indices. The weighting factor $\mathbf{z}$ have significant values for the weights with small indices and negligible values for the weight with high indices. Some of the features learned by the asymmetric network are more important than the others in representing the images. The correlation between the average weight factor $\mathbf{z}$ and the node index is 0.8420 for the asymmetric network, which indicates the features are sorted in the order of decreasing importance. In contrast, the correlation for the baseline network is 0.2833.

\begin{figure}[t!]
	\centering		
	\begin{minipage}{0.5\linewidth}		
		\centering
		{\includegraphics[width=\linewidth]{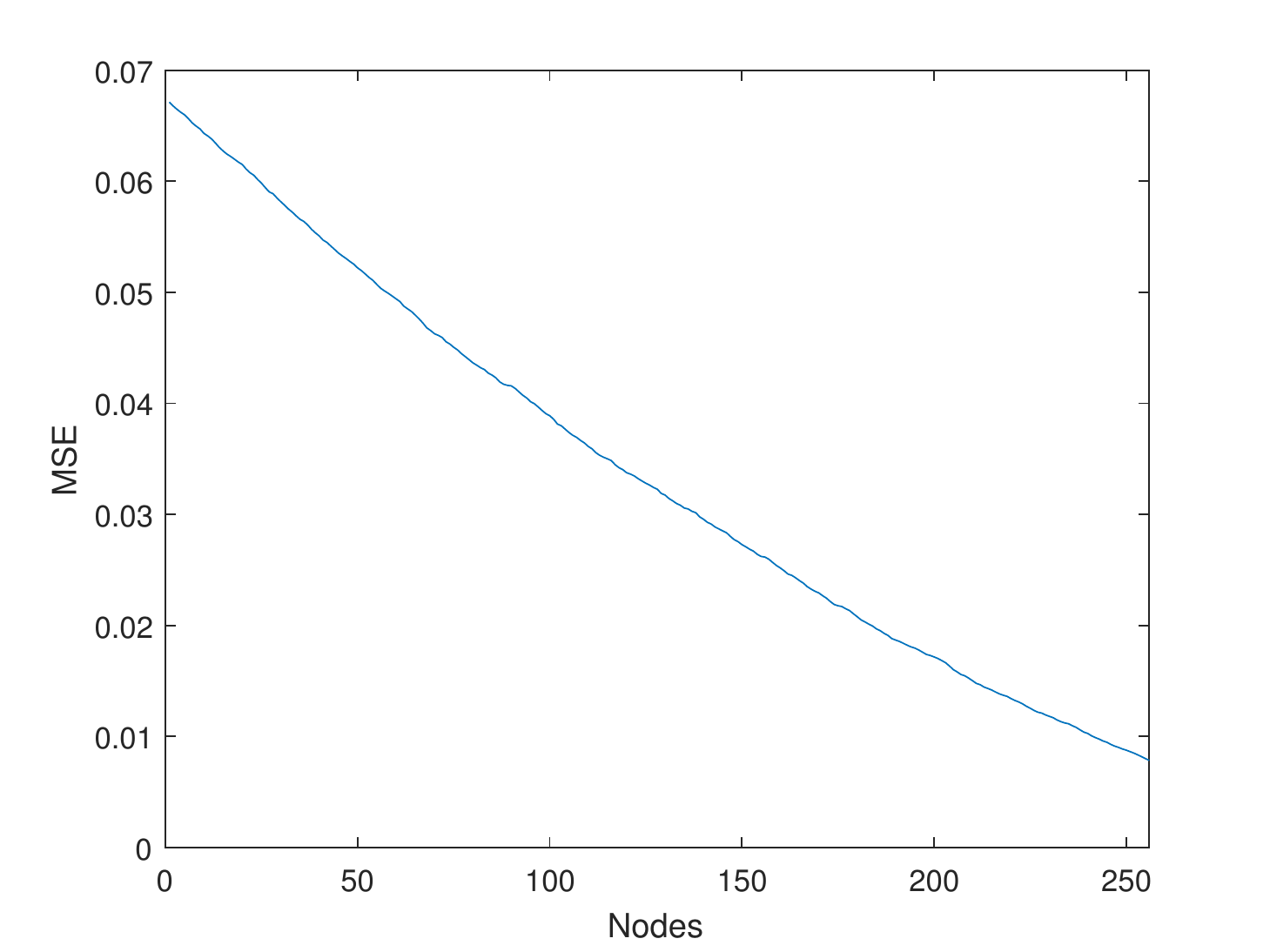}}%
		
		{\footnotesize (a)}
	\end{minipage}%
	\begin{minipage}{0.5\linewidth}		
		\centering
		{\includegraphics[width=\linewidth]{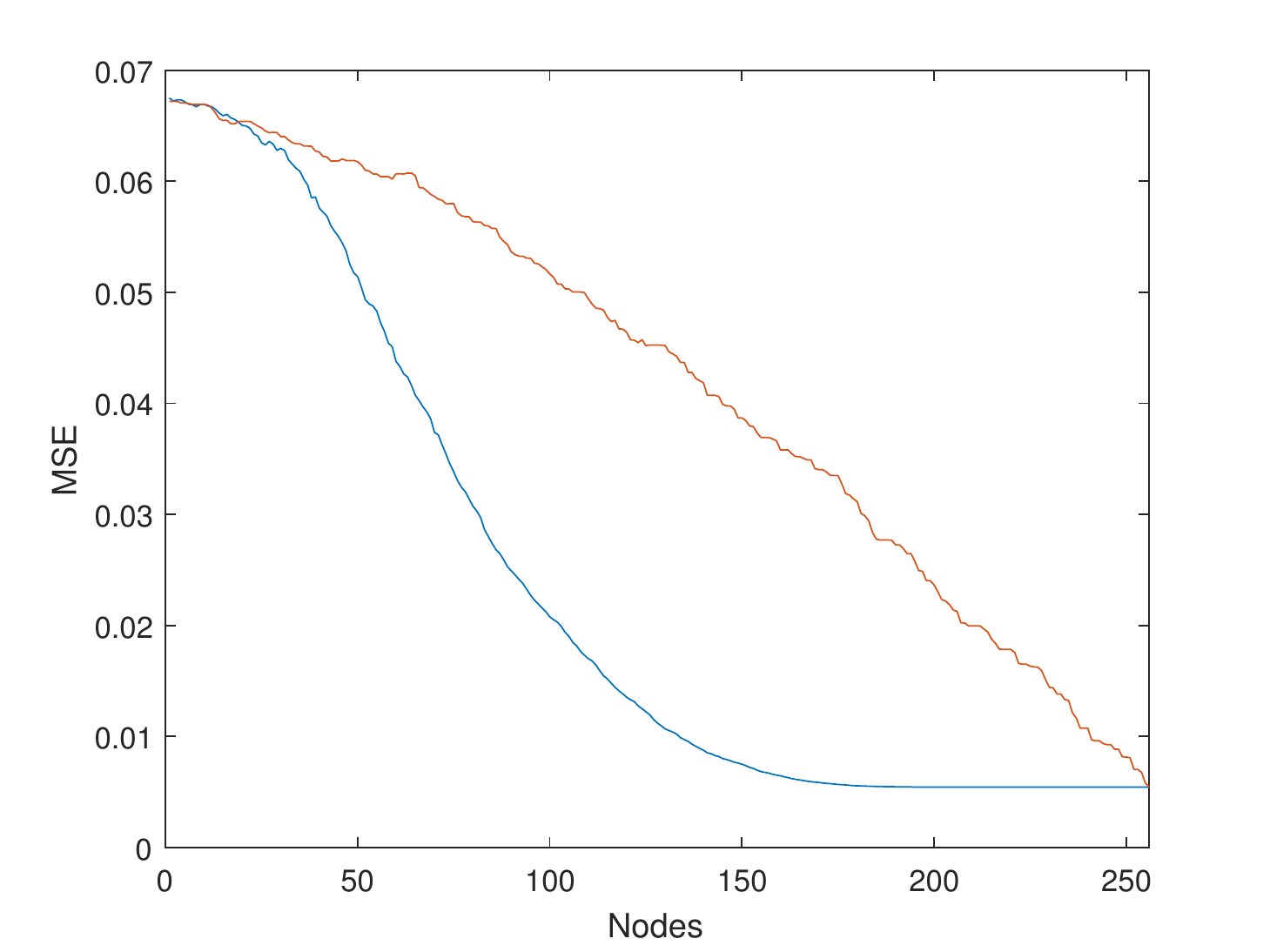}}%
		
		{\footnotesize (b)}
	\end{minipage}%
		
	\caption{MSE of images reconstructed with the first $p$ nodes with MNIST dataset,
	(a) baseline network, and
	(b) asymmetric network.
	Blue: with the first $p$ nodes according to the node indices;
	Red: with $p$ randomly selected nodes.
	}
	\label{fig:MNISTPCA}
\end{figure}

Fig. \ref{fig:MNISTPCA} shows the result of principal component analysis (PCA) on the images in the MNIST dataset, where the average mean square error (MSE) values between the inputs and their reconstructions using the first $p$ node are shown. For the asymmetric network, the average MSE drops by larger amounts when nodes with smaller indices are removed and does not improve much when adding nodes with larger indices. The average MSE of the images reconstructed by the first $p$ nodes shows that the features of the asymmetric network are sorted in order of importance. In contrast, the features of the baseline networks are of equal importance. The removal of any node results in about the same drop in the average MSE.

The feature sorting exhibited by asymmetric networks may be affected by random initialization of weights during the training. To check whether the feature-sorting is repeatable, we trained the asymmetric network five times using the same training set. To check whether it is reproducible, we also trained the asymmetric network five times, but used a different training set each time. The average pairwise mean square error (MSE) of the weights in each layer was measured and the results are shown in Table \ref{tab:R&R}. The weights learned by the asymmetric network using the same or different training sets are close together, indicating that the feature-sorting property of asymmetric networks is both repeatable and reproducible. The average pairwise MSE values of the weights of the symmetric version of the network are listed for comparison. 

\begin{table}[t!]
	\centering
	\caption{Repeatability and Reproducibility of Feature Sorting by Asymmetric Network. 
	Average Pairwise MSE of Weights.}
	\label{tab:R&R}
	\begin{tabular}{cl|c|c|c|c}
		\hline
		& & \multicolumn{2}{c|}{Repeatability} & \multicolumn{2}{c}{Reproducibility} \\ \cline{3-6}
		\multicolumn{2}{c|}{layer} & Symmetric & Asymmetric & Symmetric & Asymmetric \\ \hline
		1 & dense & 0.0017 &  0.0001& 1.698  & 0.0102\\
		2 & dense & 0.0191 & 0.0040 & 0.115 & 0.0094 \\ \hline
	\end{tabular}
\end{table}

\subsection{Deep Asymmetric Network with CIFAR-10 Dataset}
\label{sec:exdan}

We apply a set of node-wise variant activation functions to deep networks, whose performance is then analyized in object recognition using the CIFAR10 dataset \cite{krizhevsky2009learning}. The network for object recognition consists of four convolutional layers and two fully connected dense layers. All the layers utilize max pooling and activation functions. The convolution layers utilize $3\times 3$ filters. There are 128, 128, 256, and 256 nodes in the convolutional layers and 512 and 10 in the fully connected dense layers. The final layer provides the classification. The activation functions in each layer consist of a set of parameterized ReLU functions. We used the parameters shown by the $s_i$ values in Fig. \ref{fig:parameters} and applied the same learning rate that would be used for a symmetric network. Specifically, the learning rate was set to 1e-3 with a decay rate of 1e-6. Adam \cite{kingma2014adam} is used as an optimizer. The network was trained using 50,000 images and 10,000 images were used for testing.

\begin{figure}[t!]
	\centering		
	\begin{minipage}{0.5\linewidth}		
		\centering
		{\includegraphics[width=\linewidth]{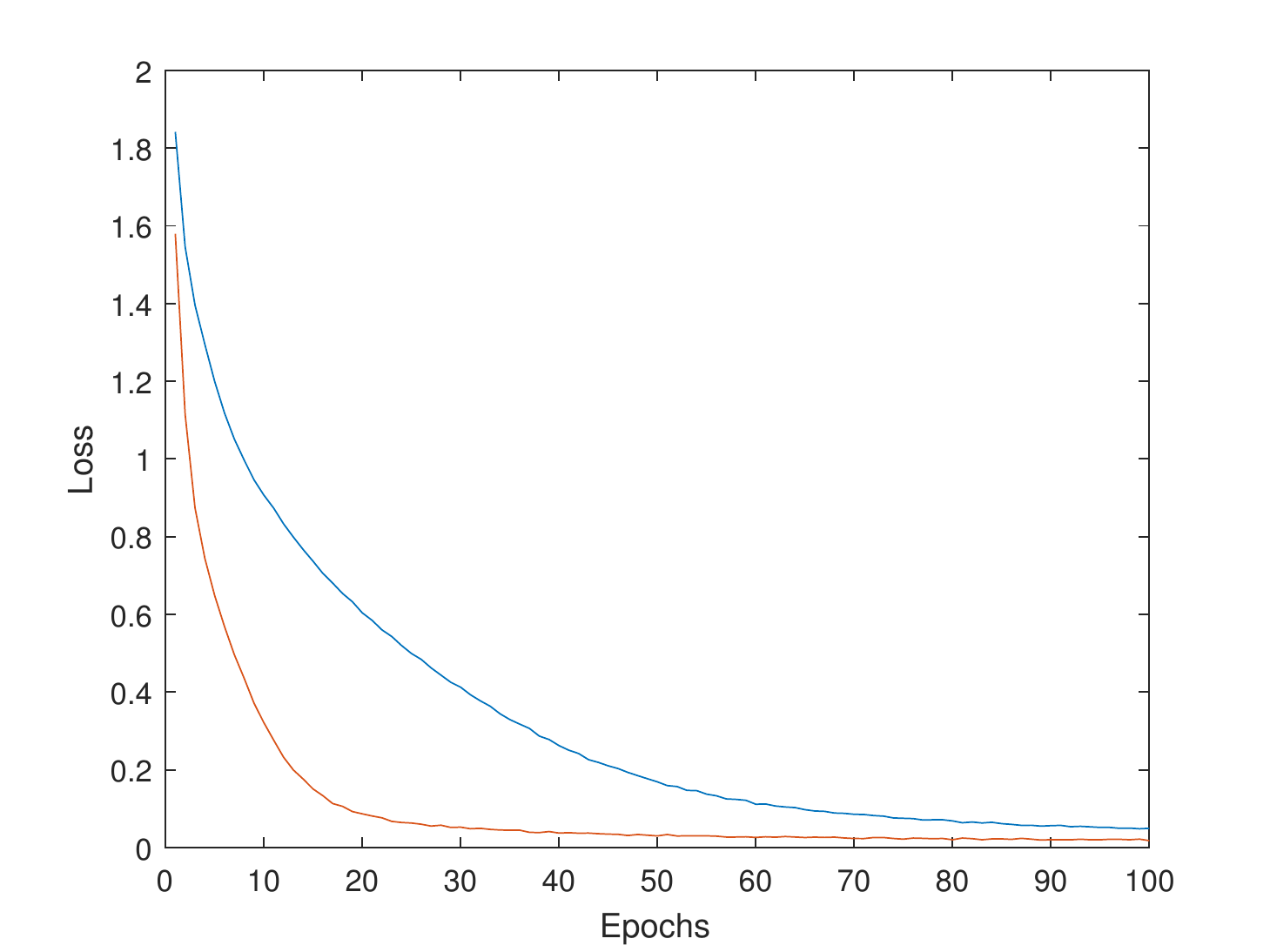}}%
		
	\end{minipage}%
	\begin{minipage}{0.5\linewidth}		
		\centering
		{\includegraphics[width=\linewidth]{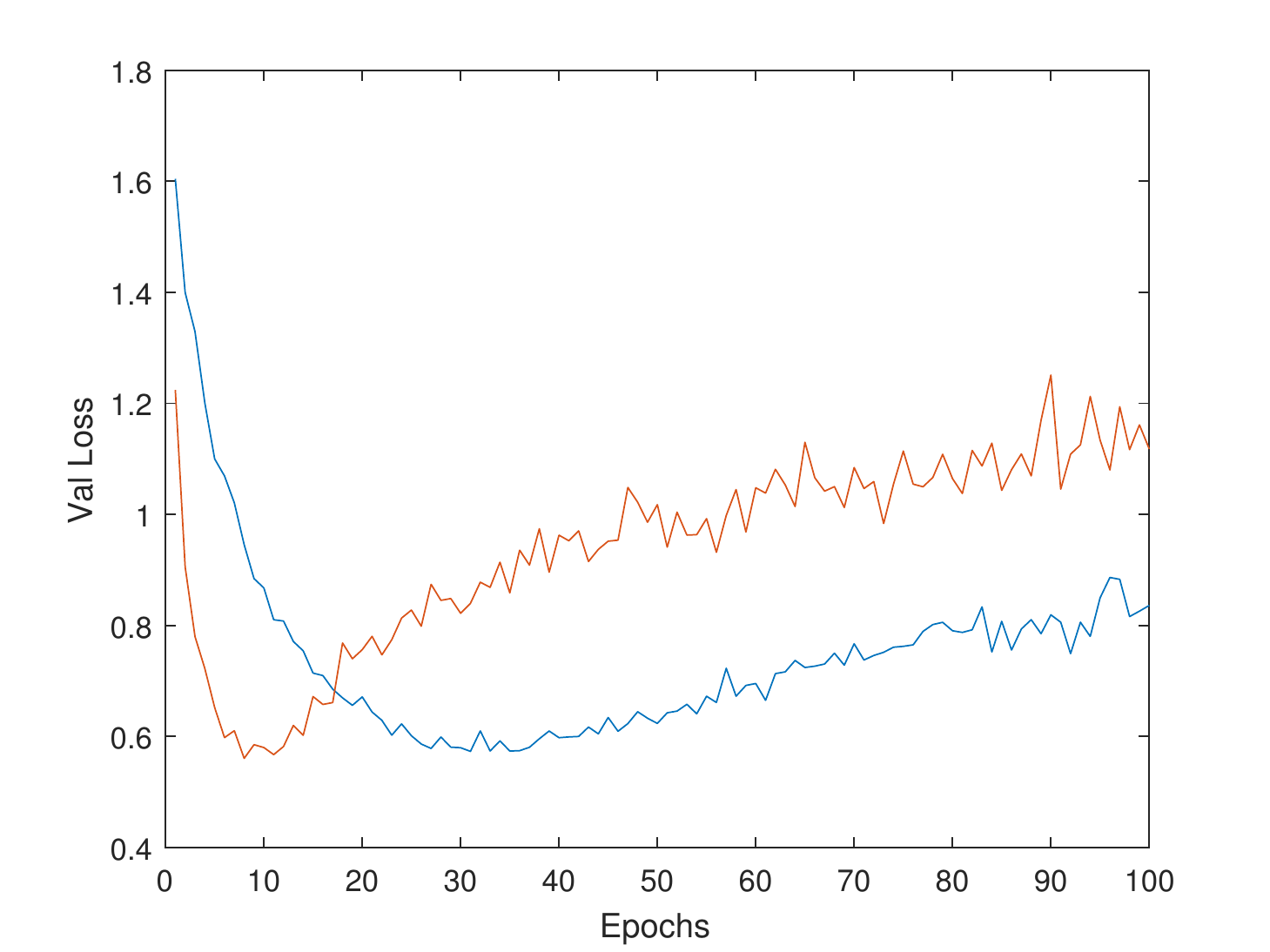}}%
		
	\end{minipage}%

	{\footnotesize (a)}
		
	\begin{minipage}{0.5\linewidth}		
		\centering
		{\includegraphics[width=\linewidth]{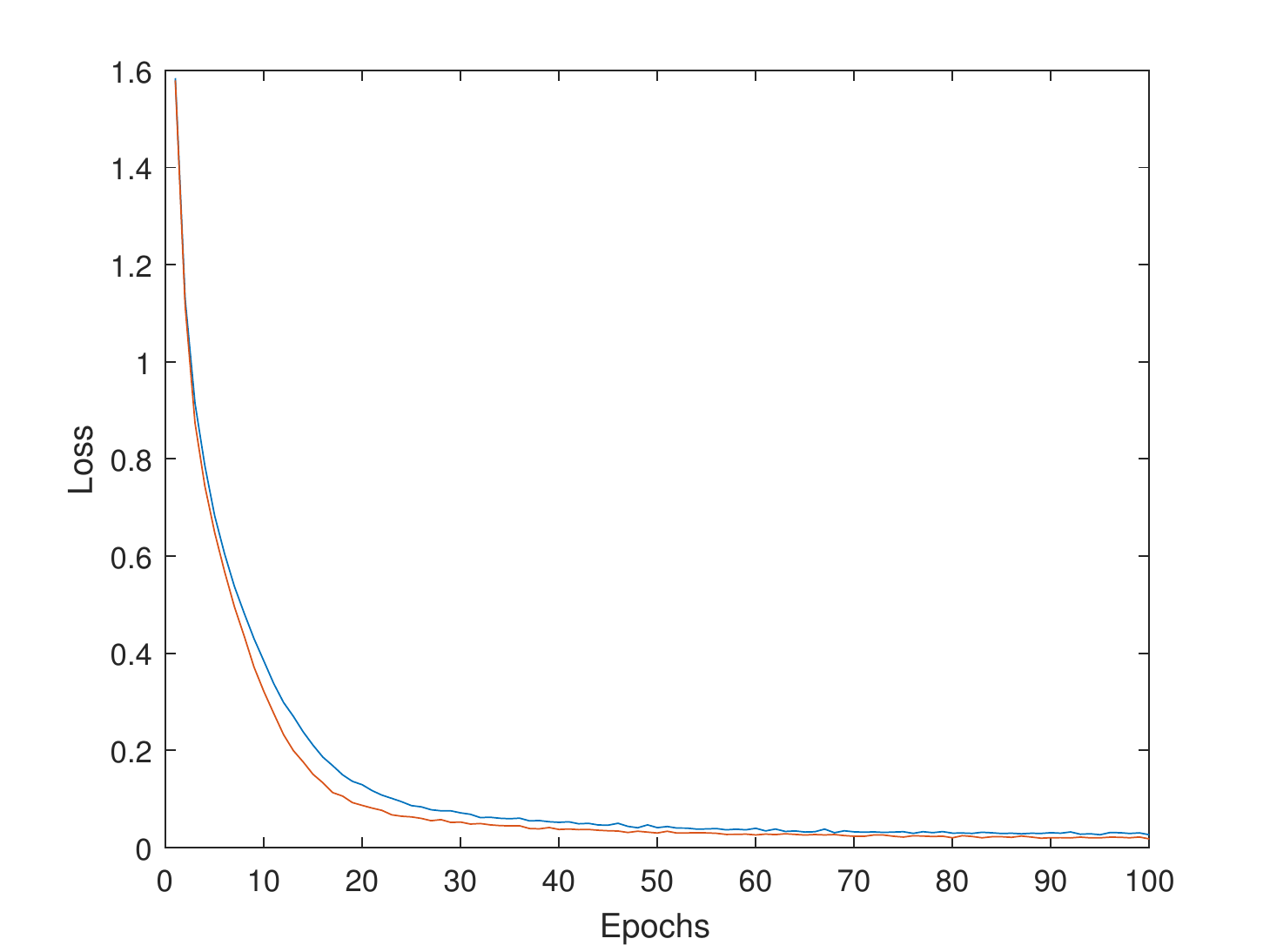}}%
	\end{minipage}%
	\begin{minipage}{0.5\linewidth}		
		\centering
		{\includegraphics[width=\linewidth]{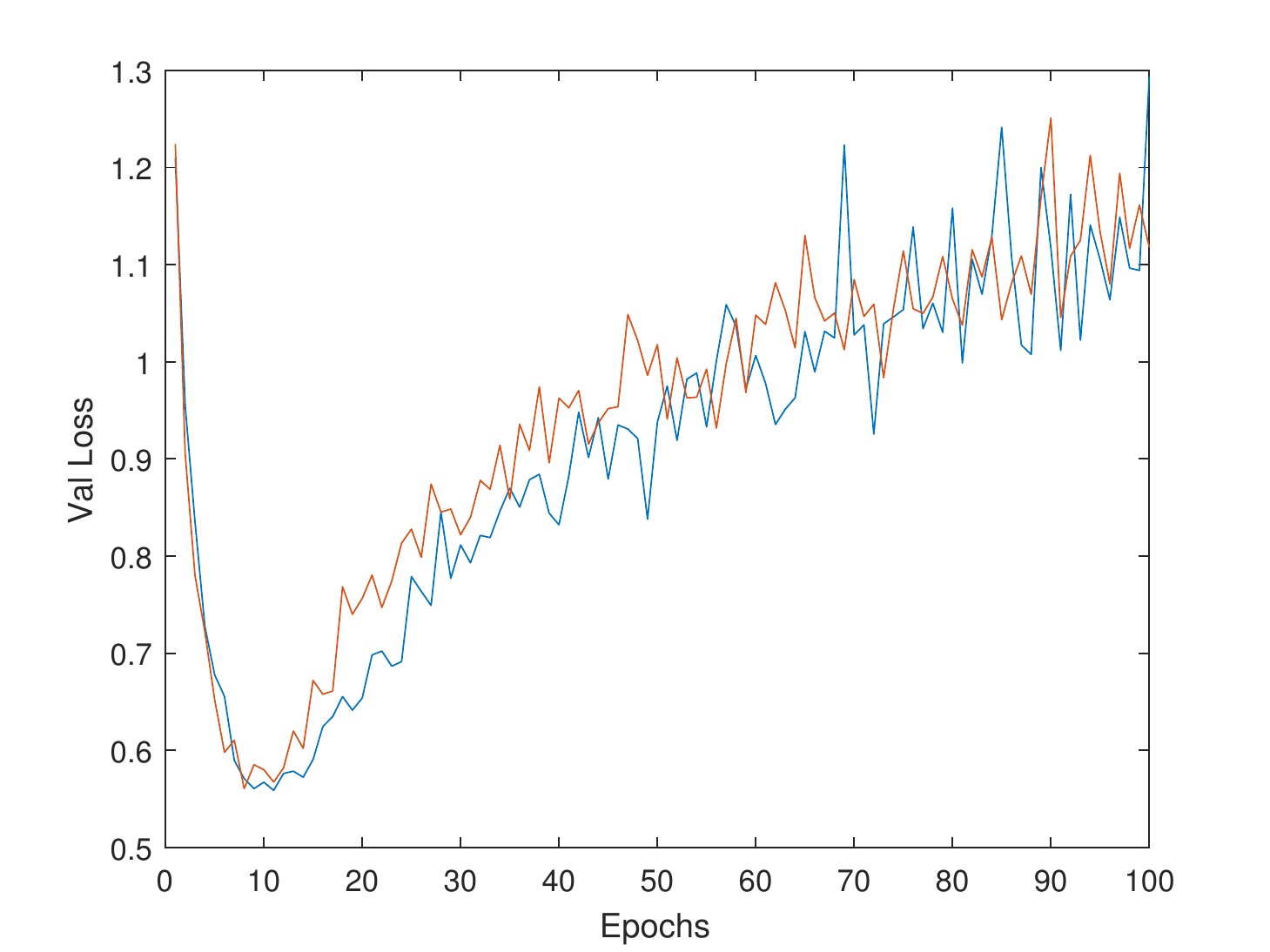}}%
	\end{minipage}%

	{\footnotesize (b)}

	\caption{Training of symmetric networks and deep asymmetric network with variant activation functions on the CIFAR-10 dataset: (a) using the same learning rate and (b) using a higher learning rate to compensate for the average slopes of the activation functions.
Left: training loss; Right: validation loss; Blue: asymmetric network; Red: symmetric network.
	}
	\label{fig:cifar10training}
\end{figure}

Fig \ref{fig:cifar10training} illustrates the training process of deep networks with the CIFAR-10 dataset. The losses and validation accuracies shown are for the asymmetric network, which uses the set of node-wise variant activation functions, and for a symmetric network that shares the same architecture but uses the same ReLU activation function for all nodes. We trained both the asymmetric and the symmetric network. With the sensitivity parameters less than one, convergence in some directions is intentionally delayed compared to convergence in other directions. The result is that the asymmetric network converges more slowly than the symmetric network as shown in Fig \ref{fig:cifar10training}(a) and (b). The training delay can be compensated for by considering the average slopes of the activation functions. The average node slopes with the parameters set shown as in Fig. \ref{fig:parameters} (b) is approximately 1/3. Therefore, we set the learning rate of the asymmetric network to three times that of the  symmetric network; then, the asymmetric network converged successfully in approximately the same number of iterations as the symmetric network as shown in (c) and (d). After training, the test accuracy was 0.8328 for the symmetric network and 0.8299 for the asymmetric network.

\begin{figure}[t!]
	\centering		
	\begin{minipage}{0.5\linewidth}		
		\centering
		{\includegraphics[width=\linewidth]{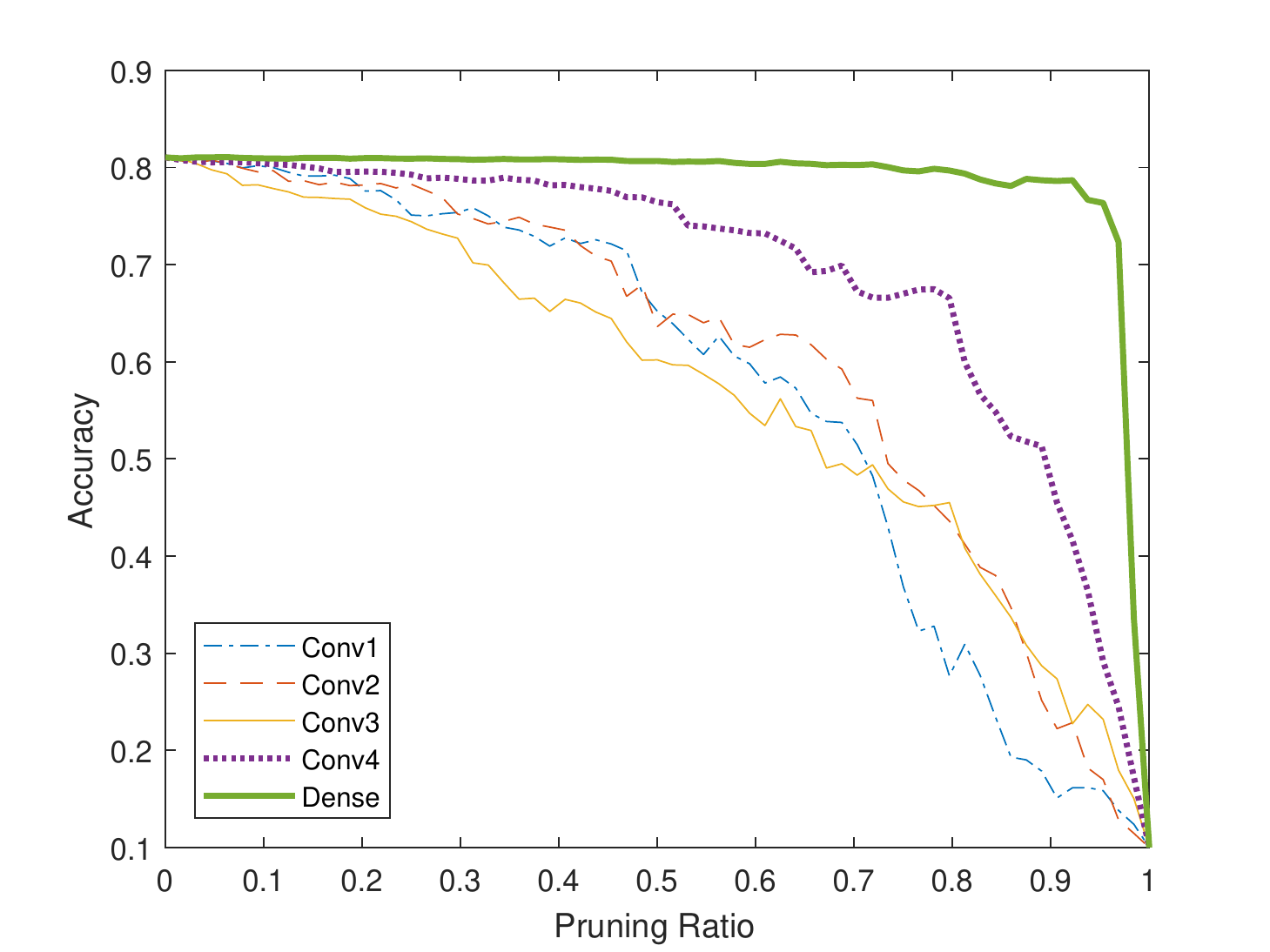}}%
		
		{\footnotesize (a)}
	\end{minipage}%
	\begin{minipage}{0.5\linewidth}		
		\centering
		{\includegraphics[width=\linewidth]{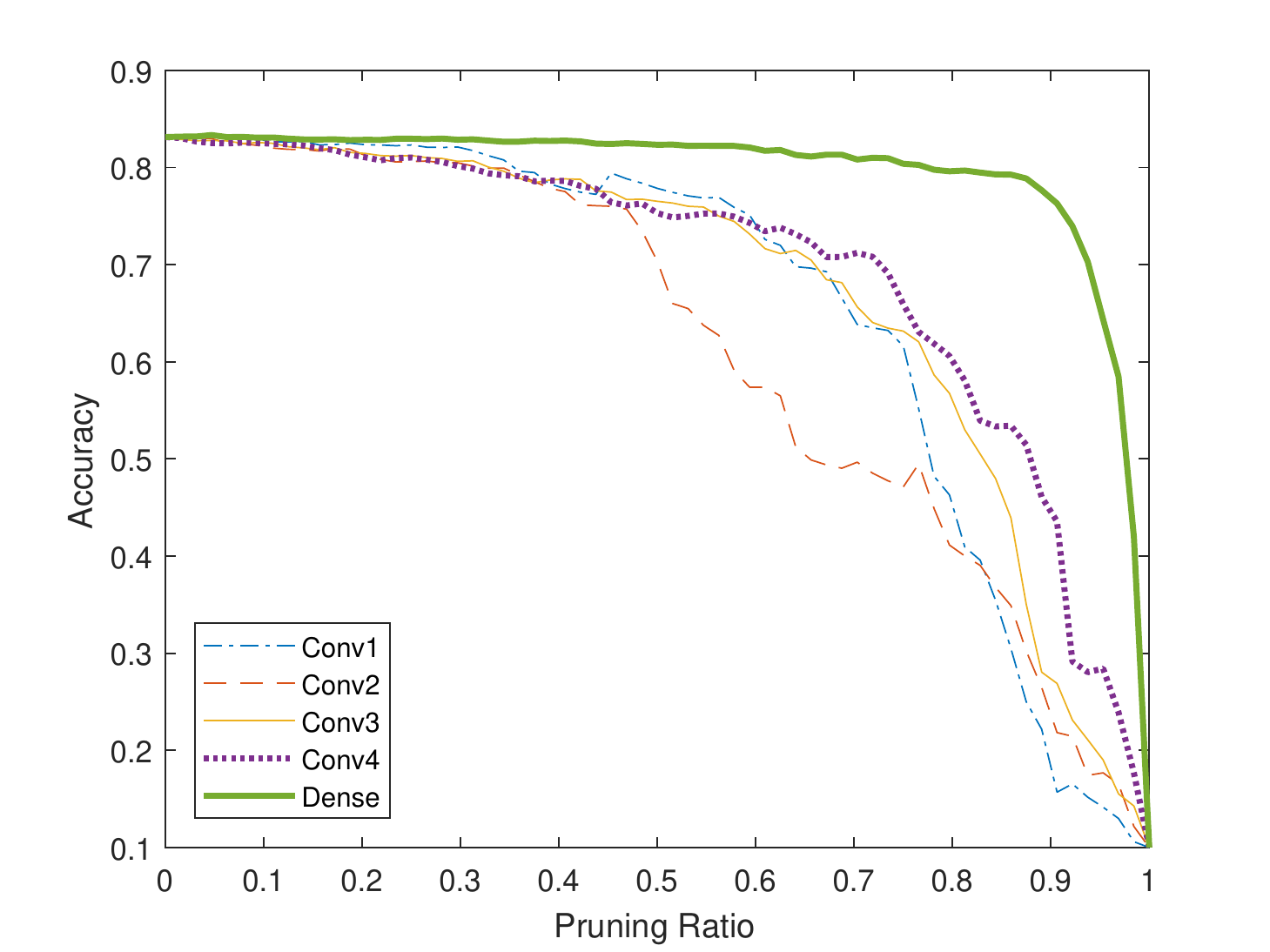}}%
		
		{\footnotesize (b)}
	\end{minipage}%
	
	\begin{minipage}{0.5\linewidth}		
		\centering
		{\includegraphics[width=\linewidth]{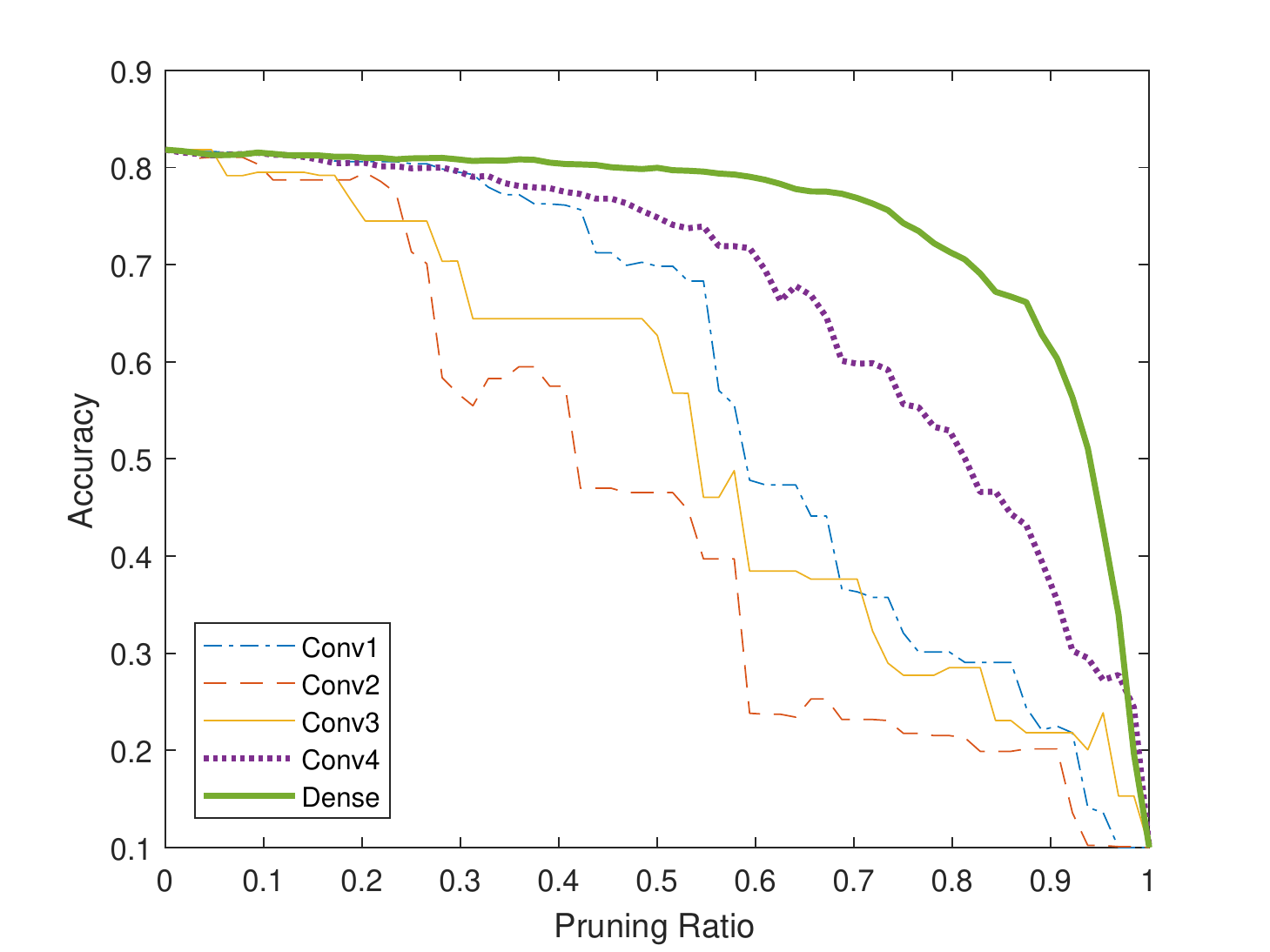}}%
		
		{\footnotesize (c)}
	\end{minipage}%
	\begin{minipage}{0.5\linewidth}		
		\centering
		{\includegraphics[width=\linewidth]{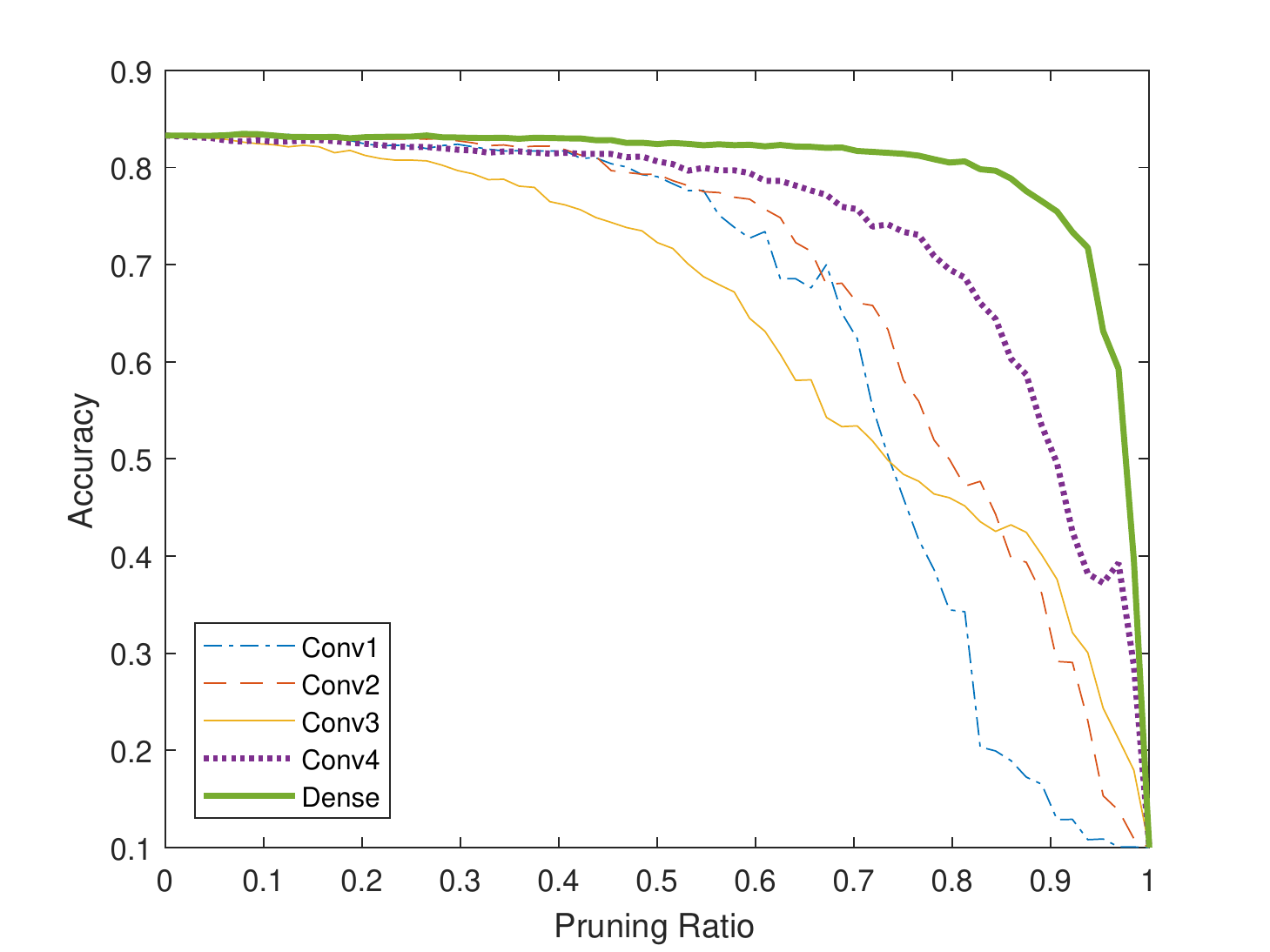}}%
		
		{\footnotesize (d)}
	\end{minipage}%
	
	\begin{minipage}{0.5\linewidth}		
		\centering
		{\includegraphics[width=\linewidth]{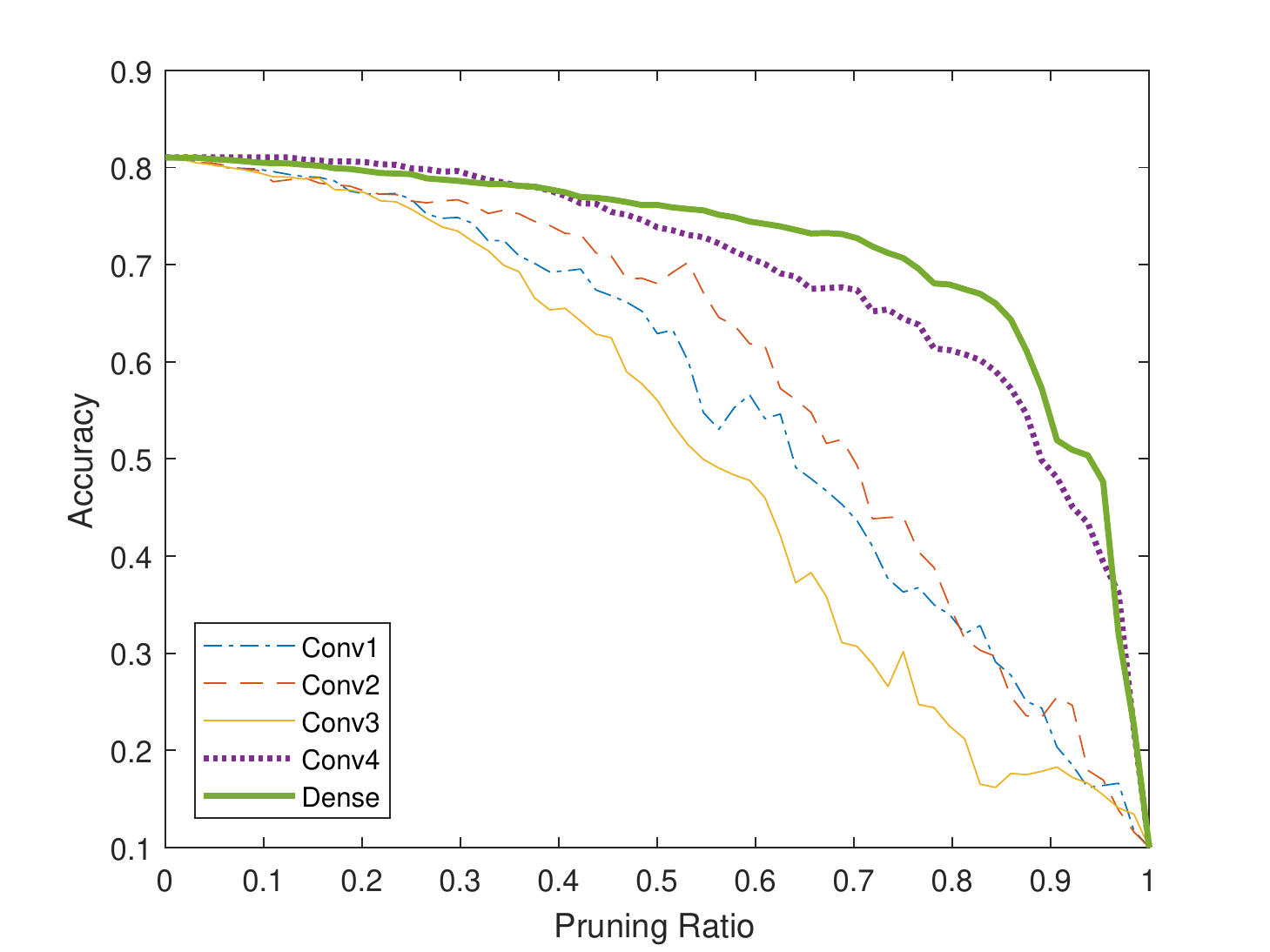}}%
		
		{\footnotesize (e)}
	\end{minipage}%
	\begin{minipage}{0.5\linewidth}		
		\centering
		{\includegraphics[width=\linewidth]{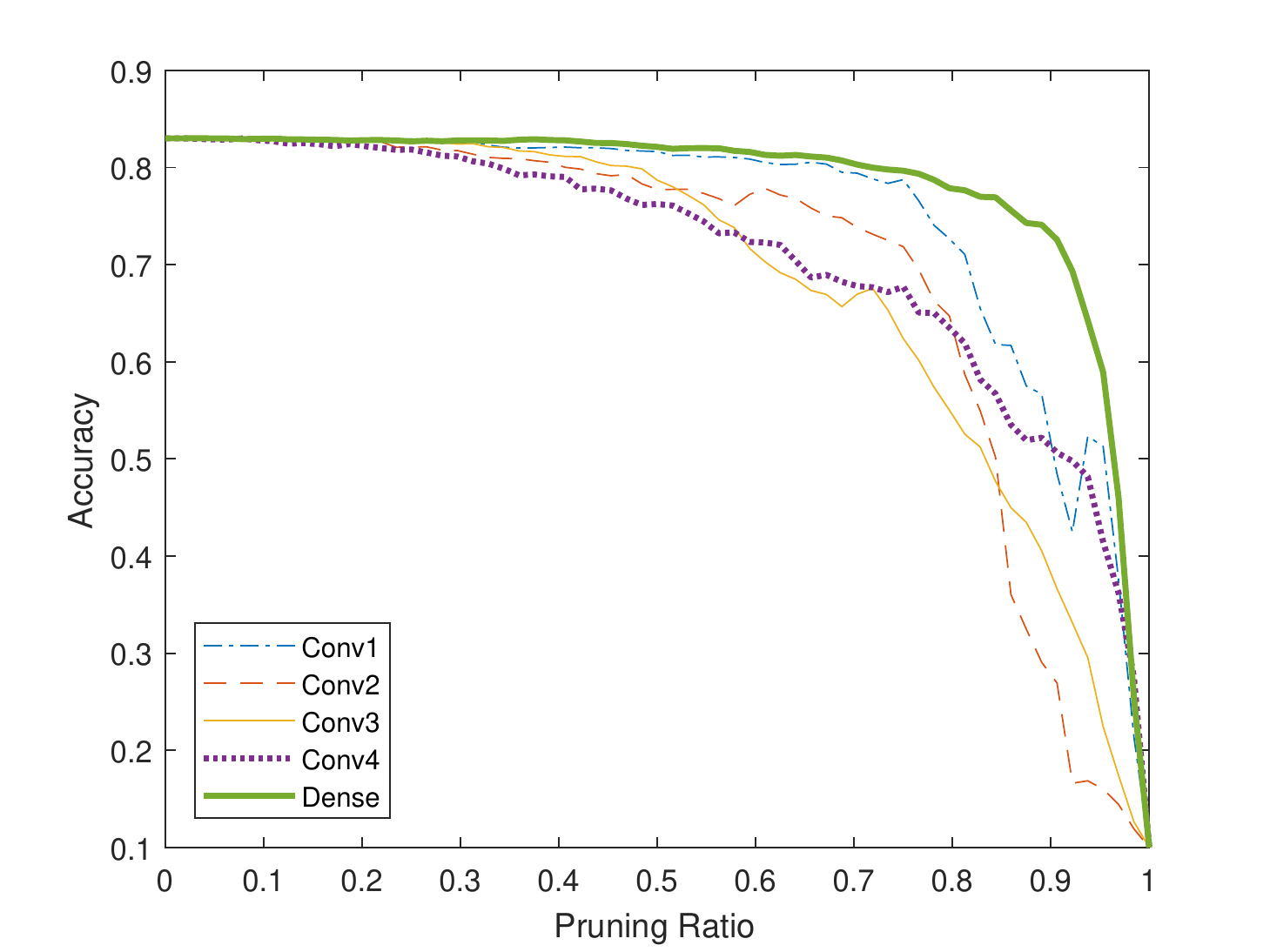}}%
		
		{\footnotesize (f)}
	\end{minipage}%
			
	\caption{Network performance when nodes are independently pruned: test accuracy with 10,000 images from the CIFAR-10 dataset. Symmetric network with (a) random pruning; (b) $l_2$ pruning; (c) $l_1$ pruning; (d) correlation pruning; (e) perturbation analysis, and (f) asymmetric network pruned from last to first nodes. 
	}
	\label{fig:cifar10sorting}
\end{figure}

Fig. \ref{fig:cifar10sorting} (f) shows how the performance of the asymmetric network deteriorates as nodes are removed from the network. The test accuracy is measured when a given percentage of nodes are removed from last to first in a layer while leaving the remaining layers intact. The test accuracy is averaged over five trainings of the network. It can be seen that the network performance gradually falls as the nodes in each layer are removed from last to first. The gradual decline of performance indicates that the nodes with larger indices are of lesser importance. For comparison, symmetric networks with the same layers and one activation function assigned to all the nodes were trained with either the $l_2$ or $l_1$ norms of the weights as regularization terms. We evaluated the performance of the symmetric network while removing nodes randomly, based on sorting by the $l_2$ or $l_1$ norms of the weights \cite{han2015learning, ishikawa1996structural}. The results are reported in Fig. \ref{fig:cifar10sorting}(a) to (c). A symmetric network with the weight correlation as a regularization term is also prepared, from which nodes are removed based on the correlation. The result is shown in Fig. \ref{fig:cifar10sorting}(d). This network represents the approaches to make the weights of networks orthogonal or independent \cite{rodriguez2016regularizing, huang2018orthogonal, bansal2018can, zhu2018improving}. Lastly, a symmetric network is prepared, from which nodes are removed based on the perturbation analysis \cite{lecun1990optimal}. The result is shown in Fig. \ref{fig:cifar10sorting}(e). The asymmetric network retains its performance better than does the symmetric network when the same number of nodes are removed from each network. For example, when we removed 90\% of the nodes in the first layer, the accuracy of the networks becomes 15.10\%, 16.85\%, 18.21\%, 19.91\%, 20.34\%, and 53.25\%, for the random, $l_2$, $l_1$, correlation, and perturbation analysis, and the proposed method, respectively. This suggests that the asymmetric network learns the importance of features better than is possible by evaluating the weights in the symmetric network using various measures. The comparison to the symmetric network trained with the correlation regularization shows that importance of the nodes with orthogonal features still have to be determined. The comparison to the perturbation analysis shows that the perturbation from the minima does not provide as accurate information on the importance of nodes. The performance improvement of the asymmetric network shows that the importance of nodes learned by the network can be used to determine which nodes to be kept or pruned.

The trained asymmetric deep network is pruned by the proposed pruning algorithm. The target accuracy used in Algorithm \ref{alg:pruning} is set to 90\% of the training accuracy of the unpruned network. The pruned asymmetric network is retrained using the training set. The ratio of the number of weights is
\begin{equation}
	\hbox{ratio} = \frac{\hbox{\# of weights after pruning}}{\hbox{\# of weights before pruning}}
\end{equation}
and the accuracy of the networks before pruning, after pruning, and after retraining are reported in Table \ref{tab:cifar10pruningaccuracy}. We were able to retrain the pruned asymmetric network without a loss of accuracy. Table \ref{tab:cifar10pruningaccuracy} also compares the accuracy when the sensitivity parameters $s_i$ of the activation functions are chosen differently from the ones shown in Fig. \ref{fig:parameters} (a), (b), and (c).  The parameters in (a) set the node sensitivities to vary linearly. The parameters in (b) set the node sensitivities to vary close to linearly at smaller indices and limits node to small sensitivities at larger indices. The parameters in (c) set the node sensitivities to one and zero for the small and large indices, respectively, and to vary linearly in the middle. The parameters in (b) yielded the highest accuracy with the smallest number of weights. Thus, we used the parameters in (b) in the rest of our experiments.  

\begin{table}[t!]
	\centering
	\caption{Performances of Pruned Asymmetric Networks with Various Choices of Activation Function Parameters as Shown in Fig. \ref{fig:parameters}, CNN with CIFAR-10 dataset.}
	\label{tab:cifar10pruningaccuracy}
	\begin{tabular}{ll|c|c|c}
		\hline
		\multicolumn{2}{l|}{Parameters} & (a)& (b) & (c) \\ \hline
		Ratio & (\# of weights) & 26.48\% & 18.13\% & 18.38\% \\ \hline
		Accuracy & Before pruning & 83.32\% & 83.34\% & 83.29\% \\
		 & After pruning & 75.00\% & 75.01\% & 74.97\% \\
		 & After retraining & 83.18\% & 83.59\% & 83.11\% \\ \hline
	\end{tabular}
\end{table}

Table \ref{tab:cifar10summary} shows the number of nodes in each layer of the asymmetric network before and after the pruning. Overall, all but 18.13\% of nodes can be removed from the network without losing accuracy. The ratio of nodes that can be removed from the layers is higher in the deeper layers. Because the computational complexity of the deeper layers is higher, pruning the deeper layers contributes more toward reducing the overall computational complexity. 

\begin{table}[t!]
	\centering
	\caption{Complexity of Pruned Asymmetric Network: CNN with CIFAR-10 dataset}
	\label{tab:cifar10summary}
	\begin{tabular}{cl|r|r|r}
		\hline
		 & & 	 \multicolumn{3}{c}{\# of weights} \\ \cline{3-5}
		\multicolumn{2}{c|}{Layer} & Before & After & Ratio \\ \hline
		1 & conv &  3584& 2165 & 60.16\%\\
		2 & conv &  147584 & 70788 & 47.96\%\\
		3 & conv &  295168& 164501 & 55.73\%\\
		4 & conv &  590080& 248248 & 42.07\%\\
		5 & dense & 3276800& 296450 & 9.05\%\\ 
		6 & dense & 5120& 770 & 15.04\%\\ \hline
		\multicolumn{2}{c|}{overall} & 4318336& 782913&  18.13 \%\\ \hline
	\end{tabular}
\end{table}

\subsection{LeNet with MNIST Dataset}
\label{sec:lenet}

\begin{figure}[t!]
	\centering		
	\begin{minipage}{0.5\linewidth}		
		\centering
		{\includegraphics[width=\linewidth]{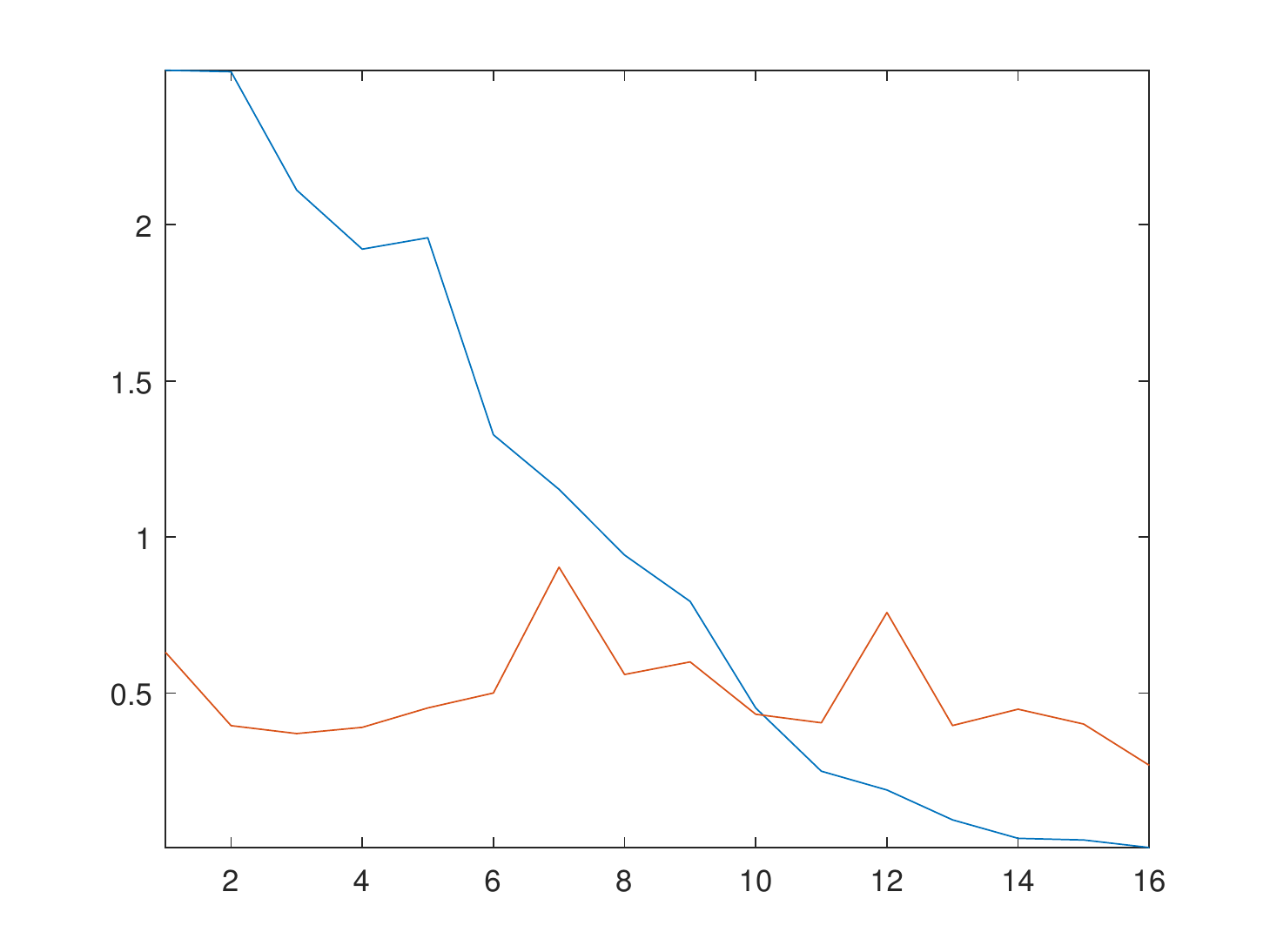}}%
		
		{\footnotesize (a)}
	\end{minipage}%
	\begin{minipage}{0.5\linewidth}		
		\centering
		{\includegraphics[width=\linewidth]{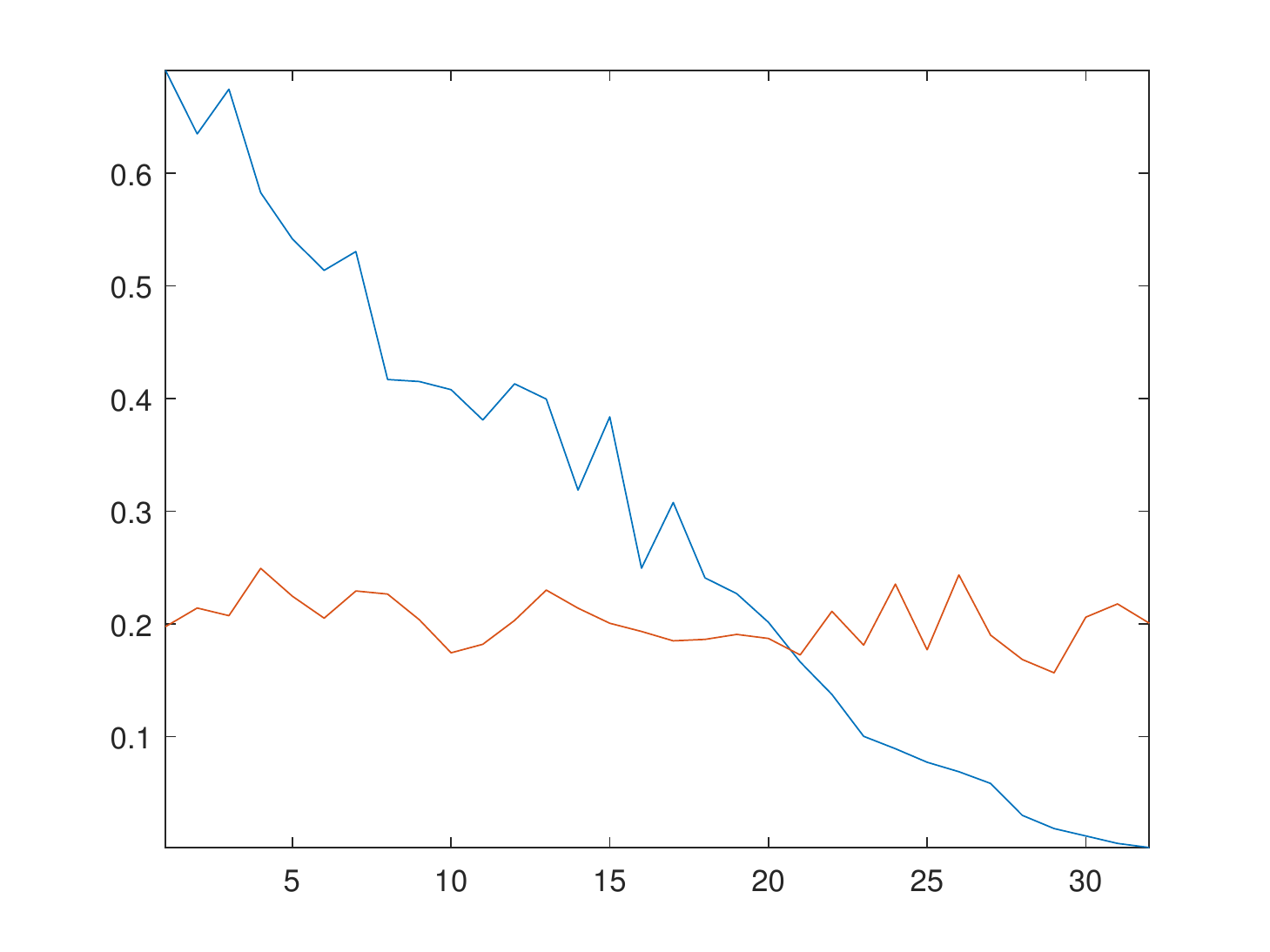}}%
		
		{\footnotesize (b)}
	\end{minipage}%
	
	\begin{minipage}{0.5\linewidth}		
		\centering
		{\includegraphics[width=\linewidth]{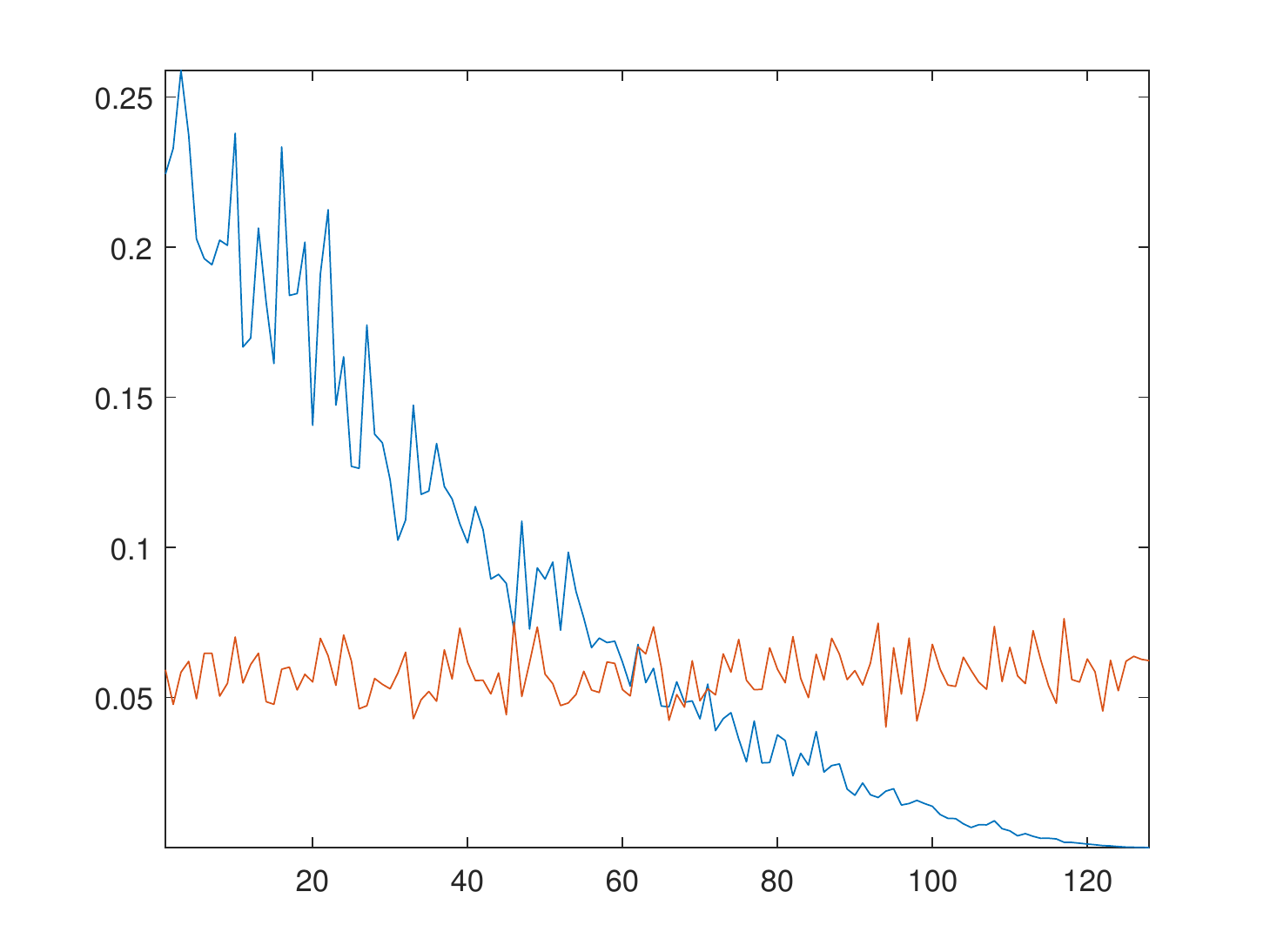}}%
		
		{\footnotesize (c)}
	\end{minipage}%
	\begin{minipage}{0.5\linewidth}		
		\centering
		{\includegraphics[width=\linewidth]{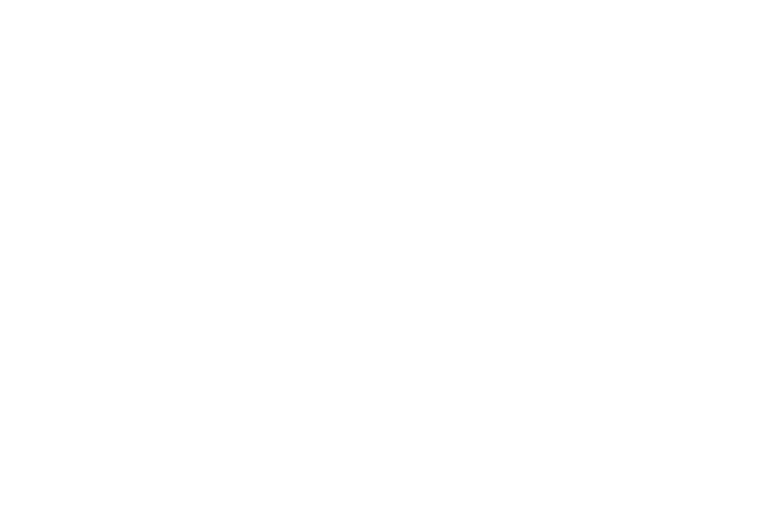}}%
		
	\end{minipage}%
			
	\caption{Validation loss when a node in a layer is removed from the network, LeNet with the MNIST dataset:
	(a) 1st convolutional layer;
	(b) 2nd convolutional layer;
	(f) 1st dense layer. 
	Red: baseline symmetric network;
	Blue: asymmetric network.
	}
	\label{fig:damage}
\end{figure}

The node-wise variant activation functions are applied to LeNet \cite{krizhevsky2012imagenet} and trained using the MNIST dataset. 
In order to validate the features are sorted in the order of importance, we measure the validation loss while removing nodes in the trained network one at a time. Fig. \ref{fig:damage} shows the validation loss of the baseline symmetric and asymmetric LeNets when a node in a layer is removed. The average validation losses of ten trainings are shown. The increases of the loss when nodes with smaller indices are removed from the network are larger for the asymmetric network. By contrast, about the same amount of increases occurred by removing of any nodes in the baseline network. The same trends are observed for all the convolutional and dense layers. The correlation between the loss and indices are given in Table \ref{tab:corr}. The loss occurred by removing a node from the asymmetric network and the node indices show high correlation, which indicates nodes in the deep asymmetric network is sorted in the order of importance.

\begin{table}[t!]
	\centering
	\caption{Correlation between the loss and the node indices: CNN with CIFAR-10 dataset}
	\label{tab:corr}
	\begin{tabular}{c|c|c}
		\hline
		layer & symmetric & asymmetric \\ \hline
		1st conv. & 0.7377& 0.9947 \\
		2nd conv. & 0.7721 & 0.9874 \\
		1st dense & 0.7322 & 0.9920 \\ \hline
	\end{tabular}
\end{table}

We compared the complexity and performance of pruned asymmetric networks to the pruning results reported in \cite{han2015learning, zhou2016less, yang2015deep, lebedev2016fast, srinivas2015data}. An asymmetric network with the same architecture as LeNet-5 \cite{krizhevsky2012imagenet} was prepared, and the weights are initialized randomly. The network was trained with MNIST dataset, pruned, and then retrained. The ratios of the weights after pruning and the classification errors are reported in Table \ref{tab:compMNIST}. Our pruning algorithm works with all the layers in LeNet-5 while achieving comparable accuracy and yielding smaller networks.

\begin{table}[t!]
	\centering
	\caption{Comparison to Pruning Methods Reported with MNIST Dataset
	}
	\label{tab:compMNIST}
	\begin{tabular}{cccrr}
		\hline 
		Method & Network & cf & Ratio& Error \\ \hline
		\cite{han2015learning} & LeNet-5 & & & 0.80\% \\
			&   & & 8.24\% & 0.77\% \\ \hline
		\cite{zhou2016less} & LeNet-5 &  & & 0.73\% \\
			& (modified) & & 10.25\% & 0.76\%\\ \hline
		\cite{yang2015deep} & LeNet-5 &  & & 0.87\% \\
			&   & & 9.01\% & 0.71\% \\ \hline
		\cite{lebedev2016fast} & LeNet-5 &   & & N/A \\
			&  & conv layers only  & 8.33\% & 1.70\% \\ \hline
		\cite{srinivas2015data} & LeNet-5 &  & & 0.94\% \\
			&   & dense layer only &  16.00\% & 1.65\% \\ 
			& & dense layer only & 12.00\%  & 2.01\% \\ \hline
		Proposed & LeNet-5 &  & &  0.81\%\\
			&   & & 6.73\% & 0.71\%\\
		\hline
	\end{tabular}
\end{table}

\subsection{Transferred VGG and ResNet with CK+ Dataset}
\label{sec:ck+}

A network can share its architecture with that of a famous network previously proven to be excellent for some other purpose. The network can be initialized using the weights transferred from the famous network and then fine-tuned with a training set prepared for a given application. Famous networks are usually trained for tasks that are more complicated than the complexity of a specific task. Hence, although weight transfer is a well-established way to achieve good network performance, the end result can be an excessively large and heavy network architecture that is computationally too expensive for a given task. In this section, we transfer famous networks to asymmetric networks and then prune the asymmetric network to design efficient and compact networks. Since weights of famous networks are already trained for good performance, we trained the asymmetric network without the correlation regularization.

We transferred VGG \cite{simonyan2014very} and ResNet \cite{he2016deep} to asymmetric networks for the purpose of facial expression recognition. The asymmetric networks consist of the same layers as VGG-16 or ResNet-50 followed by an output layer that classifies input facial images into seven emotions:
\begin{equation}
	\{\hbox{anger, contempt, disgust, fear, surprise, happiness, sadness}\}.
	\nonumber
\end{equation} 
The network is trained using the CK+ dataset \cite{lucey2010extended}. We selected 325 sequences of 118 subjects that are classified as displaying one of the seven emotions. The so-called ``apex frames'' that occur at the peak of the expression were collected as labeled facial images. The network was trained using ten-fold cross validation. The labeled images are divided into ten folds, nine of which were  used for training, and the remaining fold was used for evaluation. 

To transfer weights from a symmetric network to an asymmetric network, node sorting may be required to determine which nodes in the symmetric network should be assigned with smaller indices in the asymmetric network. However, node pre-sorting complicates the performance analysis because any performance variations may be due to either the pre-sorting or to the feature sorting of the asymmetric network. Hence, we transferred the weights without any pre-sorting of the node indices. After the weight transfer, the asymmetric network is trained with a training set. Then, the trained asymmetric network is pruned using Algorithm \ref{alg:pruning}. After the pruning, we retrain the pruned networks with the training set. 

Table \ref{tab:famuspruningsummary} shows the pruning results of the asymmetric network transferred from VGG-16 and ResNet-50. The target accuracy was set to 90\% of the training accuracy before pruning. We were able to prune the VGG and ResNet transferred asymmetric networks down to 24.48\% and 16.82\%, respectively, without loss of accuracy. The layer-wise pruning ratios are reported in Table \ref{tab:vggpruningnodes} and \ref{tab:resnetpruningnodes} for the VGG-16 and ResNet-50 transferred asymmetric networks, respectively. It is interesting to note that all the nodes in the 44, 45, 48, and 49th layers of ResNet-50 were removed during pruning. ResNet has bypass paths that enables the network to learn the error instead of the input itself. The removal of all the nodes in these convolutional layers changes the residual block operations to simple bypasses. Hence, our pruning procedure changes the network architecture from residual blocks to  convolutional layers with $1\times 1$ filters.  

\begin{table}[t!]
	\centering
	\caption{Performance of Pruned Asymmetric Network Transferred from VGG-16 and ResNet-50 on CK+ Dataset}
	\label{tab:famuspruningsummary}
	\begin{tabular}{ll|c|c}
		\hline
		\multicolumn{2}{l|}{Parameters} & VGG-16 & ResNet-50 \\ \hline
		Ratio & & 24.48\% & 16.82\% \\ \hline
		Accuracy & Before pruning & 96.30\% & 96.61\% \\
		 & After pruning & 86.70\% & 86.95\% \\
		 & After retraining & 97.51\% & 97.23\% \\ \hline
	\end{tabular}
\end{table}

\begin{table}[t!]
	\centering
	\caption{Complexity of Pruned Asymmetric Network Transferred from VGG-16 with CK+ Dataset.}
	\label{tab:vggpruningnodes}
	\begin{tabular}{cl|r|r|r}
		\hline
		 & & \multicolumn{3}{c}{\# of weights} \\ \cline{3-5}
		\multicolumn{2}{c|}{Layer} &  Before & After & Ratio \\ \hline
		1 & conv & 1792 & 1436 & 81.16\%\\
		2 & conv & 36928	& 24614 & 66.65\% \\
		3 & conv & 73856	& 48952 & 66.28 \%  \\
		4 & conv & 147584 & 98241 & 33.17\%  \\
		5 & conv & 295168 & 201990 & 68.43\%  \\
		6 & conv & 590080 & 395990 & 67.11\%  \\
		7 & conv & 590080 & 335860 & 56.92\% \\
		8 & conv & 1180160 & 564910 & 47.87\% \\
		9 & conv & 2359808 & 845684 & 35.84\%  \\
		10 & conv & 2359808 & 532610 & 22.57\%  \\
		11 & conv & 2359808 & 498460 & 21.12\%  \\
		12 & conv & 2359808 & 351360 & 14.89\%  \\
		13 & conv & 2359808 & 189430 & 8.03\% \\
		14 & dense & 2097152 & 27152 & 1.29\% \\
		15 & dense & 7168 & 392 & 5.48\%  \\ \hline
		\multicolumn{2}{c|}{overall} & 16819008 &  4117082 & 24.48\% \\ \hline
	\end{tabular}
\end{table}

\begin{table}[t!]
	\centering
	\caption{Complexity of Pruned Asymmetric Network Transferred from ResNet-50 with CK+ Dataset. }
	\label{tab:resnetpruningnodes}
	\begin{tabular}{cl|r|r|r}
		\hline
		 & & \multicolumn{3}{c}{\# of weights} \\ \cline{3-5}
		\multicolumn{2}{c|}{Layer} &  Before & After & Ratio \\ \hline
1	&	conv	&	9472	&	7948	&	83.91\%	\\
2	&	conv	&	4160	&	2481	&	59.63\%	\\
3	&	conv	&	36928	&	2052	&	5.56\%	\\
4	&	conv	&	16640	&	12979	&	78.00\%	\\
5	&	conv	&	16640	&	16640	&	100.00\%	\\
6	&	conv	&	16448	&	12876	&	78.28\%	\\
7	&	conv	&	36928	&	19783	&	53.57\%	\\
8	&	conv	&	16640	&	11443	&	68.77\%	\\
9	&	conv	&	16448	&	9535	&	57.97\%	\\
10	&	conv	&	36928	&	11985	&	32.46\%	\\
11	&	conv	&	16640	&	8346	&	50.15\%	\\
12	&	conv	&	32896	&	24004	&	72.97\%	\\
13	&	conv	&	147584	&	66100	&	44.79\%	\\
14	&	conv	&	66048	&	38502	&	58.29\%	\\
15	&	conv	&	131584	&	131584	&	100.00\%	\\
16	&	conv	&	65664	&	40784	&	62.11\%	\\
17	&	conv	&	147584	&	57492	&	38.96\%	\\
18	&	conv	&	66048	&	40550	&	61.39\%	\\
19	&	conv	&	65664	&	33807	&	51.48\%	\\
20	&	conv	&	147584	&	37026	&	25.09\%	\\
21	&	conv	&	66048	&	24422	&	36.98\%	\\
22	&	conv	&	65664	&	25291	&	38.52\%	\\
23	&	conv	&	147584	&	26843	&	18.19\%	\\
24	&	conv	&	66048	&	21862	&	33.10\%	\\
25	&	conv	&	131328	&	59098	&	45.00\%	\\
26	&	conv	&	590080	&	95373	&	16.16\%	\\
27	&	conv	&	263168	&	67994	&	25.84\%	\\
28	&	conv	&	525312	&	525312	&	100.00\%	\\
29	&	conv	&	262400	&	13120	&	5.00\%	\\
30	&	conv	&	590080	&	25135	&	4.26\%	\\
31	&	conv	&	263168	&	23347	&	8.87\%	\\
32	&	conv	&	262400	&	7893	&	3.01\%	\\
33	&	conv	&	590080	&	4442	&	0.75\%	\\
34	&	conv	&	263168	&	7578	&	2.88\%	\\
35	&	conv	&	262400	&	6560	&	2.50\%	\\
36	&	conv	&	590080	&	8136	&	1.38\%	\\
37	&	conv	&	263168	&	15462	&	5.88\%	\\
38	&	conv	&	262400	&	19680	&	7.50\%	\\
39	&	conv	&	590080	&	26627	&	4.51\%	\\
40	&	conv	&	263168	&	16794	&	6.38\%	\\
41	&	conv	&	262400	&	9225	&	3.52\%	\\
42	&	conv	&	590080	&	2109	&	0.36\%	\\
43	&	conv	&	263168	&	3686	&	1.40\%	\\
44	&	conv	&	524800	&	0	&	0.00\%	\\
45	&	conv	&	2359808	&	0	&	0.00\%	\\
46	&	conv	&	1050624	&	2048	&	0.19\%	\\
47	&	conv	&	2099200	&	2099200	&	100.00\%	\\
48	&	conv	&	1049088	&	0	&	0.00\%	\\
49	&	conv	&	2359808	&	0	&	0.00\%	\\
50	&	conv	&	1050624	&	1050624	&	100.00\%	\\
51	&	conv	&	1049088	&	31555	&	3.01\%	\\
52	&	conv	&	2359808	&	3606	&	0.15\%	\\
53	&	conv	&	1050624	&	7373	&	0.70\%	\\
54	&	dense	&	8388608	&	543950	&	6.48\%	\\
55	&	dense	&	7168	&	465	&	6.48\%	\\ \hline
\multicolumn{2}{c|}{overall}	&	31877248	&	5360724	&	16.82\%	\\ \hline
	\end{tabular}
\end{table}

\subsection{Transferred CNN with NTU RGB+D Action Recognition Dataset}
\label{sec:NTU}

We apply the asymmetric network to action recognition using the NTU RGB+D action recognition dataset \cite{du2015skeleton}. The NTU RGB+D action recognition dataset consists of 56,880 video sequences of 60 types of actions of 40 people captured from three camera views. The performance of the network is evaluated using the cross-subject protocol. The video sequences of 20 people are used for the training and those of the other 20 people are used for the test.

The network for action recognition is from \cite{du2015skeleton}. The network uses two stream CNN, with three convolutional layers in each stream, followed by four dense layers. We initialized the asymmetric network with the same weights provided with the code. The activation functions in each layer are assigned with a set of parameterized ReLU functions. To assign the parameter $s_i$ to a node, node sorting may be required to determine which nodes in the symmetric network should be assigned with smaller indices in the asymmetric network. We can sort the nodes of a symmetric network using various measure, for example, by the $l_2$ norm of the weights, and then assign the node indices based on the sorted results. However, such node pre-sorting complicates the performance analysis because any performance variations may be due to either the pre-sorting or to the feature sorting of the asymmetric network. Hence, we assign the parameters without any pre-sorting of the node indices. After the parameter assignment, the asymmetric network is trained with NTU RGB+D action recognition dataset. Since the initial weights are already trained for good performance, we trained the asymmetric network without the correlation regularization. 

The trained asymmetric deep network is pruned by the proposed pruning algorithm. The target accuracy used in Algorithm \ref{alg:pruning} is set to 90\% of the training accuracy of the unpruned network. The pruned asymmetric network is retrained using the training set. Table \ref{tab:ntusummary} shows the number of nodes in each layer of the asymmetric network before and after the pruning. Overall, all but 38.63\% of nodes can be removed from the network without losing accuracy. Accuracy of the network at each step of pruning is given in Table \ref{tab:ntupruningaccuracy}. 

\begin{table}[t!]
	\centering
	\caption{Complexity of Pruned Asymmetric Network: Deep Network in \cite{du2015skeleton} with NTU RGB+D Action Recognition Dataset.}
	\label{tab:ntusummary}
	\begin{tabular}{ccl|r|r|r}
		\hline
		& & & 	 \multicolumn{3}{c}{\# of weights} \\ \cline{4-6}
		\multicolumn{3}{c|}{Layer} & Before & After & Ratio \\ \hline
		 \multicolumn{3}{c|}{transform} &  750 & 750 & 100.00\%\\ \hline
		1 & & conv &  1728 & 1728 & 100.00\%\\
		3 & & conv &  73728 & 70272 & 95.31\%\\
		5 && conv &  294912 & 126270 & 42.82\%\\ \hline
		& 2 & conv &  1728 & 1728 & 100.00 \%\\
		& 4 & conv &  73728 & 47808 & 64.84\%\\
		& 6 & conv &  294912 & 57519 & 19.50\%\\ \hline
		 \multicolumn{2}{c}{7} & dense & 524544& 157645 & 30.05\%\\ 
		 \multicolumn{2}{c}{8} & dense & 32896 & 26368 & 80.16\%\\
		 \multicolumn{2}{c}{9} & dense & 12384 & 12384 & 100.00\%\\ 
		 \multicolumn{2}{c}{10} & dense & 5820 & 5820 & 100.00\%\\ \hline
		\multicolumn{3}{c|}{overall} & 1317880 & 509042 &  38.63 \%\\ \hline
	\end{tabular}
\end{table}

\begin{table}[t!]
	\centering
	\caption{Performances of Pruned Asymmetric Network, Deep Network in \cite{du2015skeleton} with NTU RGB+D Action Recognition Dataset.}
	\label{tab:ntupruningaccuracy}
	\begin{tabular}{ll|c}
		\hline
		Ratio & (\# of weights) & 38.63\%  \\ \hline
		Accuracy & Before pruning & 80.66\% \\
		 & After pruning & 66.75\% \\
		 & After retraining & 80.24\%  \\ \hline
	\end{tabular}
\end{table}

\subsection{Transferred VGG and ResNet with CIFAR-10 Dataset}
\label{sec:CIFAR}

We compared the complexity and performance of the pruned asymmetric networks with the pruning results reported in \cite{li2016pruning}. VGG-16 and ResNet-56 are transferred to asymmetric networks, trained on the CIFAR-10 dataset, pruned, and retrained. The weight ratios after pruning and the classification errors are reported in Table \ref{tab:compCIFAR}. Our pruning algorithm works with all the layers in ResNet. The proposed method reports comparable accuracy and yields smaller networks.

\begin{table}[t!]
	\centering
	\caption{Comparison to Pruning Methods Reported with CIFAR-10 Dataset}
	\label{tab:compCIFAR}
	\begin{tabular}{cccrr}
		\hline
		Method & Network & cf & Ratio & Error \\ \hline
		\cite{li2016pruning} & VGG-16 & & & 6.75\% \\
			&   & & 36.0\% & 6.60\% \\ \hline
		Proposed & VGG-16 & & & 6.67\% \\
			&   & & 29.6\% & 6.17\% \\ \hline
		\cite{li2016pruning} & ResNet-56 &  &   & 6.96\% \\
			&   & 1st layer of residual blocks & 86.3\% & 6.94\% \\ \hline
		Proposed & ResNet-56 &   & & 6.82\% \\
			&  & & 26.2\% & 6.75\% \\
		\hline
	\end{tabular}
\end{table}

\subsection{Transferred ResNet with ImageNet}
\label{sec:ImageNet}

We also compared the complexity and performance of the pruned asymmetric networks with the pruning results reported in \cite{luo2017thinet, luo2017entropy, zhuang2018discrimination, xu2018hybrid, he2017channel}. ResNet-50 is transferred to asymmetric networks, trained on the ImageNet dataset, pruned, and retrained. The weight ratios after pruning and the classification errors are reported in Table \ref{tab:compImageNet}. Our pruning algorithm reports comparable accuracy and yields smaller networks.

\begin{table}[t!]
	\centering
	\caption{Comparison to Pruning Methods Reported with ImageNet Dataset}
	\label{tab:compImageNet}
	\begin{tabular}{cccrr}
		\hline
		Method & Network & cf & Ratio & Accuracy \\ \hline
		\cite{luo2017thinet} & ResNet-50 &  &  & 72.88\% \\
			&   & & 66.3\% & 72.04\% \\ 
			&   & & 48.4\% & 71.01\% \\ 
			&   & & 33.9\% & 68.42\% \\ \hline
		\cite{luo2017entropy} & ResNet-50 &  &  & 72.88\% \\
			&   & & 93.5\% &73.56\% \\ 
			&   & & 84.9\% & 72.89\% \\ 
			&   & & 68.0\% & 70.84\% \\ \hline
		\cite{zhuang2018discrimination} & ResNet-50 &  &  & 76.01\% \\
			&   & WM & 48.5\% &73.20\% \\ 
			&   & WM+ & 48.5\% & 72.89\% \\ 
			&   & DCP & 48.5\% & 70.84\% \\ \hline
		\cite{xu2018hybrid} & ResNet-50 &  &  & 76.01\% \\
			&   & & 67.3\% &74.87\% \\ \hline
		\cite{he2017channel} & ResNet-50 &  &  & 75.30\% \\
			&   & & 64.0\% &72.30\% \\ \hline
		Proposed & ResNet-50 & & & 75.06\% \\
			&   & & 47.0\% & 74.97\% \\ 
			&   & & 39.3\% & 74.56\% \\ \hline
	\end{tabular}
\end{table}

\section{Conclusion}
\label{sec:conclusion}

By assigning nodes unequal sensitivities through a set of node-wise variant activation functions, the resulting asymmetric networks not only learn the features of inputs but also the importance of those features. We validated the feature-sorting property of asymmetric networks through experiments using shallow, deep, and transferred asymmetric networks. Asymmetric networks can be used to learn and represent inputs accurately and efficiently with a small number of features.

\bibliographystyle{IEEEtran}
\bibliography{./a.bib}

\newpage

\begin{IEEEbiographynophoto}{Jinhyeok Jang} received a  B.S. degree in 2014 and an M.S. degree in 2016 from the
School of Electrical and Computer Engineering of the Ulsan National Institute of Science and Technology, Ulsan, South Korea. He currently works at the Electronics and Telecommunications Research Institute (ETRI), Daejeon, South Korea. His research interests include image processing, blur estimation, human facial recognition, and human action recognition.
\end{IEEEbiographynophoto}

\begin{IEEEbiographynophoto}{Hyunjoong Cho} received a B. S. degree in 2015 and an M.S. degree in
2017 from the Ulsan National Institute of Science and Technology, Ulsan,
South Korea, where he is currently pursuing a doctor’s degree in electrical
and computer engineering. His research interests include image processing,
human perception, and human recognition.
\end{IEEEbiographynophoto}

\begin{IEEEbiographynophoto}{Jaehong Kim} received his Ph.D. degree in computer engineering from Kyungpook National University, Daegu, Rep. of Korea, in
1996. He has worked as a researcher at ETRI since 2001. His research interests include eldercare robotics and social HRI frameworks.
\end{IEEEbiographynophoto}

\begin{IEEEbiographynophoto}{Jaeyeon Lee} received his Ph.D. degree from Tokai University, Japan, in 1996. Since 1986, he has worked as a research scientist at the Electronics and Telecommunications Research Institute, Daejeon, Rep. of Korea. His research interests include robotics, pattern recognition, computer vision, and biometric security
\end{IEEEbiographynophoto}

\begin{IEEEbiographynophoto}{Seungjoon Yang } (S'99-M'00) 
received a B.S. degree from Seoul National University, Seoul, Korea in 1990 and M.S. and Ph.D. degrees from the University of Wisconsin-Madison in 1993 and 2000, respectively, all in electrical engineering. He worked at the Digital Media R\&D Center at Samsung Electronics Co., Ltd. from September 2000 to August 2008 and is currently with the School of Electrical and Computer Engineering at the Ulsan National Institute of Science and Technology in Ulsan, Korea. His research interests include image processing, estimation theory, and multi-rate systems.
Professor Yang received the Samsung Award for the Best Technology Achievement of the Year in 2008 for his work on  the premium digital television platform project.
\end{IEEEbiographynophoto}

\vfill

\end{document}